\documentclass{article}

\usepackage{microtype}
\usepackage{graphicx}
\usepackage{subfigure}
\usepackage{booktabs} 

\usepackage{hyperref}

\usepackage[accepted]{icml2024}

\usepackage{amsmath}
\usepackage{amssymb}
\usepackage{mathtools}
\usepackage{amsthm}
\usepackage{bbm}
\usepackage{multirow}

\usepackage[capitalize,noabbrev,nameinlink]{cleveref}

\theoremstyle{plain}
\newtheorem{theorem}{Theorem}[section]
\newtheorem{proposition}[theorem]{Proposition}

\theoremstyle{definition}

\theoremstyle{remark}
\newtheorem{remark}[theorem]{Remark}

\usepackage[textsize=tiny]{todonotes}

\usepackage{acronym}
\acrodef{grsnn}[GRSNN]{Graph Reasoning Spiking Neural Network}
\acrodef{snn}[SNN]{Spiking Neural Network}
\acrodef{ann}[ANN]{Artificial Neural Network}
\acrodef{ai}[AI]{Artificial Intelligence}
\acrodef{gnn}[GNN]{Graph Neural Network}
\acrodef{mr}[MR]{Mean Rank}
\acrodef{mrr}[MRR]{Mean Reciprocal Rank}
\acrodef{ac}[AC]{Accumulate}
\acrodef{mac}[MAC]{Multiply-and-Accumulate}
\acrodef{auroc}[AUROC]{Area Under the Receiver Operating Characteristic Curve}
\acrodef{ap}[AP]{Average Precision}
\acrodef{lif}[current-based LIF]{current-based Leaky Integrate and Fire}
\acrodef{srm}[SRM]{Spike Response Model}
\acrodef{mlp}[MLP]{Multi-Layer Perceptron}
\acrodef{add}[Add]{Addition}
\acrodef{mul}[Mul]{Multiplication}

\usepackage{xspace}
\makeatletter
\DeclareRobustCommand\onedot{\futurelet\@let@token\@onedot}
\def\@onedot{\ifx\@let@token.\else.\null\fi\xspace}
\def\eg{\emph{e.g}\onedot} 

\def\ie{\emph{i.e}\onedot}

\def\wrt{w.r.t\onedot}

\makeatother

\makeatletter
\renewcommand{\paragraph}{%
  \@startsection{paragraph}{4}%
  {\z@}{0ex \@plus 0ex \@minus 0ex}{-1em}%
  {\hskip\parindent\normalfont\normalsize\bfseries}%
}
\makeatother

\crefname{algorithm}{Alg.}{Algs.}
\Crefname{algocf}{Algorithm}{Algorithms}
\crefname{section}{Section}{Sections}
\Crefname{section}{Section}{Sections}
\crefname{table}{Table}{Tables}
\Crefname{table}{Table}{Tables}
\crefname{figure}{Fig.}{Figs.}
\Crefname{figure}{Figure}{Figures}
\crefname{equation}{Eq.}{Eqs.}
\Crefname{equation}{Equation}{Equations}
\crefname{appendix}{Appendix}{Appendices}
\Crefname{appendix}{Appendix}{Appendices}

\icmltitlerunning{Temporal Spiking Neural Networks with Synaptic Delay for Graph Reasoning}

\begin{document}

\twocolumn[
\icmltitle{Temporal Spiking Neural Networks with Synaptic Delay for Graph Reasoning}



\icmlsetsymbol{equal}{*}

\begin{icmlauthorlist}
\icmlauthor{Mingqing Xiao}{pku}
\icmlauthor{Yixin Zhu}{pkuai}
\icmlauthor{Di He}{pku,pkuai}
\icmlauthor{Zhouchen Lin}{pku,pkuai,pazhou}
\end{icmlauthorlist}

\icmlaffiliation{pku}{National Key Lab of General AI, School of Intelligence Science and Technology, Peking University}
\icmlaffiliation{pkuai}{Institute for Artificial Intelligence, Peking University}
\icmlaffiliation{pazhou}{Pazhou Laboratory (Huangpu), Guangzhou, China}

\icmlcorrespondingauthor{Zhouchen Lin}{zlin@pku.edu.cn}

\icmlkeywords{Spiking Neural Networks, Spiking Time, Synaptic Delay, Graph Reasoning, Neuromorphic Computing}

\vskip 0.3in
]



\printAffiliationsAndNotice{}  

\begin{abstract}
Spiking neural networks (SNNs) are investigated as biologically inspired models of neural computation, distinguished by their computational capability and energy efficiency due to precise spiking times and sparse spikes with event-driven computation. A significant question is how SNNs can emulate human-like graph-based reasoning of concepts and relations, especially leveraging the temporal domain optimally. This paper reveals that SNNs, when amalgamated with synaptic delay and temporal coding, are proficient in executing (knowledge) graph reasoning. It is elucidated that spiking time can function as an additional dimension to encode relation properties via a neural-generalized path formulation. Empirical results highlight the efficacy of temporal delay in relation processing and showcase exemplary performance in diverse graph reasoning tasks. The spiking model is theoretically estimated to achieve $20\times$ energy savings compared to non-spiking counterparts, deepening insights into the capabilities and potential of biologically inspired SNNs for efficient reasoning. The code is available at \url{https://github.com/pkuxmq/GRSNN}.
\end{abstract}

\section{Introduction}

\begin{figure}[t!]
    \centering
    \includegraphics[width=\linewidth]{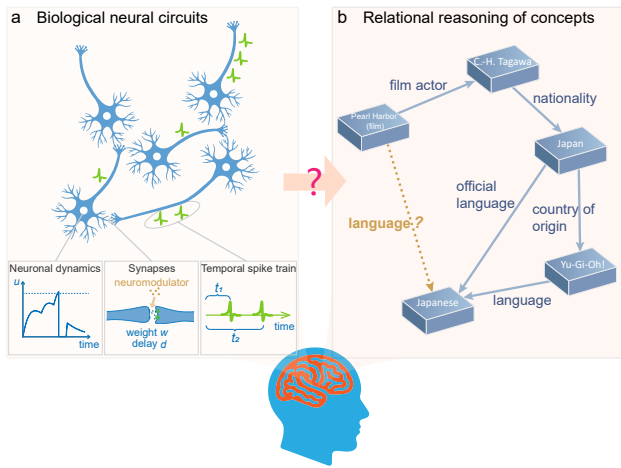}
    \vspace{-4mm}
    \caption{\textbf{Depiction of spiking neural networks and knowledge graph reasoning.} (a) A representation of biological neural circuits, showcasing spiking neurons, their inherent dynamics, synaptic interconnections, and the propagation of temporal spike trains. (b) The process of relational reasoning of concepts, exemplified through the link prediction task in knowledge graphs.}

    \label{fig:overview}
    \vspace{-2mm}
\end{figure}

\acp{snn}, inspired by the detailed dynamics of biological neurons, are recognized as more biologically plausible models for neural computation and are distinguished as the third generation of neural network models, owing to their advanced computational capabilities derived from spiking time~\citep{maass1997networks}. Unlike traditional \acp{ann}, \acp{snn} integrate neuronal dynamics using differential equations and leverage sparse spike trains in the temporal domain for information transition (\cref{fig:overview}a), enhancing the encoding of information in biological brains~\citep{reinagel2000temporal,huxter2003independent} and exhibiting increased expressive power when incorporating delay variables~\citep{maass1997networks}. The utilization of sparse, event-based computation in \acp{snn} facilitates energy-efficient operation on neuromorphic hardware with parallel in-/near-memory computing~\citep{davies2018loihi,pei2019towards,rao2022long}, making \acp{snn} increasingly prominent as powerful and efficient neuro-inspired models in \ac{ai} applications~\citep{rueckauer2017conversion,shrestha2018slayer,roy2019towards,bellec2020solution,stockl2021optimized,yin2021accuratenmi,rao2022long,xiao2022online,li2023scaling,zhang2024efficient}. Despite these advancements, critical inquiries remain unresolved regarding the solution by \acp{snn} for human-like graph-based reasoning of concepts or relations and an improved utilization of spiking time for information processing.

Symbolic and relational reasoning is a cornerstone of human intelligence and advanced \ac{ai} capabilities~\citep{kemp2008discovery,santoro2017simple,rao2022long,nickel2015review} and can often be formulated as graph reasoning with tasks like link prediction in knowledge graphs (\cref{fig:overview}b)~\citep{nickel2015review}. For example, it can be evaluated by machine learning tasks of knowledge graph completion~\citep{nickel2015review} and inductive relation prediction~\citep{yang2017differentiable,teru2020inductive}, resembling humans' ability to reason new relations between entities based on commonsense knowledge graphs or generalize relations to new analogous conditions. Investigating how underlying mechanisms of neural computation can realize this reasoning capability is pivotal for understanding human intelligence and advancing \ac{ai} systems, as graph reasoning is important for extensive \ac{ai} tasks such as knowledge graphs, recommendation systems, and drug or material design~\citep{wang2023scientific}. While various machine learning methods, including path-based~\citep{lao2010relational,yang2017differentiable,sadeghian2019drum}, embedding~\citep{bordes2013translating,yang2015embedding,sunrotate}, and \acp{gnn}~\citep{schlichtkrull2018modeling,vashishthcomposition,teru2020inductive,zhu2021neural}, have been proposed for graph reasoning tasks, the efficacy of bio-inspired models in achieving comparable performance remains largely unexplored. Existing attempts, such as entity embedding by spiking times of single neurons~\citep{dold2021spike,dold2022relational} or in-context relational reasoning~\citep{rao2022long}, have not addressed how reasoning paths can be propagated, especially with optimal utilization of temporal information at the network level, and have shown limitations in inductive generalization, interpretability, and performance in large knowledge graphs. 

Moreover, the importance of spiking time in \acp{snn}~\citep{maass1997networks,reinagel2000temporal,huxter2003independent} and its potential in \ac{ai} applications necessitate further exploration. Many previous works have primarily focused on enhancing \acp{snn} as energy-efficient alternatives to \acp{ann} for tasks like image classification~\citep{rueckauer2017conversion,shrestha2018slayer,xiao2022online}, with an emphasis on spike counts. Efforts to leverage spiking time have explored encoding information for single neurons by the time to first spike~\citep{mostafa2017supervised,comsa2020temporal,dold2021spike}, the interval between spikes~\citep{dold2022relational}, or adopting different weight coefficients at different times~\citep{stockl2021optimized}, and some have delved into temporal processing tasks like time series classification~\citep{yin2021accuratenmi,rao2022long,patino2023empirical,hammouamri2024learning}. However, more systematic utilization of synaptic delay at the network level and the coding principles embedded in neuronal spike trains are areas that warrant deeper investigation for better understanding and application of \acp{snn} in extensive \ac{ai} tasks.

In this work, we introduce \ac{grsnn}, a novel method allowing \acp{snn} to adeptly solve knowledge graph reasoning tasks by leveraging synaptic delay to encode relational information. This method enables the temporal domain of \acp{snn} to act as an additional dimension to process edge and path properties at the \emph{network} level, offering a fresh perspective on temporal information processing and coding in \acp{snn}. 

We consider link prediction tasks of knowledge graphs and 
\ac{grsnn} is proposed as a neural generalization to the path formulation of graph algorithms, drawing inspiration from existing works~\citep{aimone2021provable,zhu2021neural}. Path formulation is important to graph reasoning due to better interpretability and inductive generalization ability~\citep{zhu2021neural,yang2017differentiable,sadeghian2019drum}. 
We generalize the thought---\acp{snn} can provide a parallelizable and efficient solution to traditional graph path tasks---into \ac{ai} applications of graph reasoning. 
It can serve as a neural generalization of Dijkstra's algorithm with learnable synaptic delays representing the properties of graph edges (also coupled with synaptic weights), enabling high-performance and interpretable solutions.

Experiments on diverse graph prediction tasks are conducted to assess the effectiveness of \ac{grsnn}. The results underscore the advantage of synaptic delay in encoding relation information in \acp{snn} for competitive performance, revealing a potential mechanism of spiking neurons for knowledge reasoning, and demonstrate the efficiency of \ac{grsnn} by fewer parameters and spike computation, with a theoretical estimation indicating significant energy savings compared to non-spiking counterparts. These insights enhance our understanding of the role of neuro-inspired models in graph-based reasoning tasks, central to human intelligence, and emphasize the potential of the temporal domain of \acp{snn} in developing energy-efficient solutions for graph-based \ac{ai} applications.

\section{Preliminaries}

\subsection{Spiking Neural Networks}\label{sec:snn}

\acp{snn} are brain-inspired models comprising spiking neurons that communicate through temporal spike trains. In this work, we employ the \ac{lif} spiking neuron model, which can be equivalently represented using the \ac{srm} form. In this model, each spiking neuron maintains a membrane potential $u$, integrating input spike trains according to the dynamics:
\begin{equation}
    \small
    \tau_m\frac{du}{dt} = -(u(t)-u_{rest}) + R\cdot I(t),\quad u(t) < V_{th},
\end{equation}
where $I$ is the input current, $V_{th}$ is the threshold, $R$ is the resistance, and $\tau_m$ is the membrane time constant. When $u$ reaches $V_{th}$ at time $t^f$, a spike is emitted, and $u$ is reset to the resting potential $u=u_{rest}$, typically set to zero. The neuron's output spike train is represented as $s(t) = \sum_{t^f}\delta(t-t^f)$, using the Dirac delta function.

Neurons are interconnected through synapses with weight and delay (axonal or synaptic delay, for simplicity, we call synaptic delay in this paper). The model for input current is given by:
\begin{equation}
    \small
    \tau_c\frac{dI_i}{dt} = -I_i(t) + \sum_j w_{ij} s_j(t - d_{ij}) + b_i,
\end{equation}
where $w_{ij}$ and $d_{ij}$ are the synaptic weight and delay from neuron $j$ to neuron $i$, respectively, $b_i$ is a bias term representing background current, and $\tau_c$ is another time constant. Given the reset mechanism, the equivalent \ac{srm} form is:
\begin{equation}
    \small
    \vspace{-1mm}
    \begin{aligned}
        u_i(t) = &u_{rest}+\sum_jw_{ij}\int_0^t\kappa(\tau-d_{ij})s_j(t-\tau) \mathrm{d}\tau \\
        &+ \int_0^t\nu(\tau)s_i(t-\tau) \mathrm{d}\tau,
    \end{aligned}
    \vspace{-1mm}
\end{equation}
with $\kappa(\tau)$ being the temporal kernel function for input spikes and $\nu(\tau)=-(V_{th}-u_{rest})e^{-\frac{\tau}{\tau_m}}$ representing the reset kernel. Assuming $\tau_c=\tau_m$, the input kernel becomes $\kappa(\tau)=\frac{R}{\tau_m}\cdot \tau e^{-\frac{\tau}{\tau_m}}$ for $\tau\geq0$ and $\kappa(\tau)=0$ for $\tau<0$. Setting $R=e$, the kernel simplifies to $\kappa(\tau)=\frac{\tau}{\tau_m}e^{1-\frac{\tau}{\tau_m}}$, which is commonly used~\citep{shrestha2018slayer}. 

In practice, we simulate SNNs using the discrete computational form of the \ac{lif} model:
\begin{equation}
    \small
    \left\{
    \begin{aligned}
        &I_i[t + 1] = e^{-\frac{\Delta \tau}{\tau_m}} I_i[t] + \alpha \left(\sum_j  w_{ij} s_j[t - d_{ij}] + b_i\right),\\
        &u_i[t + 1] = e^{-\frac{\Delta \tau}{\tau_m}} u_i[t] (1 - s_i[t]) + I_i[t + 1],\\
        &s_i[t + 1] = H(u_i\left[t+1\right] - V_{th}),\\
    \end{aligned}
    \right.
\end{equation}
where $H(x)$ is the Heaviside step function, $s_i[t]$ is the spike signal at discrete time step $t$, $\Delta \tau$ is the discretization interval, and $\alpha$ denote the coefficient $\frac{e \Delta \tau}{\tau_m}$.

Utilizing the equivalent \ac{srm} formulation and surrogate derivatives for the spiking function, gradients for parameters, including $w_{ij}$ and $d_{ij}$, can be computed through backpropagation over time~\citep{shrestha2018slayer}. Specifically, the non-differentiable term $\frac{\partial s_i[t]}{\partial u_i[t]}$ is substituted by surrogate derivatives of a smooth function, such as the derivative of the sigmoid function: $\frac{\partial s}{\partial u}=\frac{1}{a_1}\frac{e^{(V_{th}-u)/a_1}}{(1+e^{(V_{th}-u)/a_1})^2}$, with $a_1$ as a hyperparameter. The gradients are then calculated as $\frac{\partial \mathcal{L}}{\partial w_{ij}}=\sum_t\frac{\partial \mathcal{L}}{\partial s_i[t]}\frac{\partial s_i[t]}{\partial u_i[t]}\frac{\partial u_i[t]}{\partial w_{ij}}$ and $\frac{\partial \mathcal{L}}{\partial d_{ij}}=\sum_t\frac{\partial \mathcal{L}}{\partial s_i[t]}\frac{\partial s_i[t]}{\partial u_i[t]}\frac{\partial u_i[t]}{\partial r_{ij}[t]}\frac{\partial r_{ij}[t]}{\partial d_{ij}}$, where $r_{ij}[t]=\sum_{\tau=0}^t \kappa(\tau-d_{ij})s_j[t-\tau]$, and $\frac{\partial r_{ij}[t]}{\partial d_{ij}}=-\sum_{\tau=0}^t \dot{\kappa}(\tau-d_{ij})s_j[t-\tau]$ ($\dot{\kappa}$ denotes the derivative of the kernel $\kappa$). In a discrete setting, $\mathbf{d}_{ij}$ should be integers, and we employ the straight-through-estimator to train a quantized real-valued variable. For additional details, please refer to \cref{supp:sec:snn_training}. In this study, we primarily focus on parameters $w_{ij}$ and $d_{ij}$, leaving the exploration of heterogeneous neurons for future work.

\subsection{Link Prediction of Graphs}

We consider link prediction tasks of (knowledge) graphs. A knowledge graph is denoted by $\mathcal{G}=(\mathcal{V}, \mathcal{E}, \mathcal{R})$, with $\mathcal{V}$, $\mathcal{E}$, and $\mathcal{R}$ representing the sets of graph nodes, graph edges, and relation types, respectively. We also consider homogeneous graphs $\mathcal{G}=(\mathcal{V}, \mathcal{E})$ as a special case with only one relation type. The task is to predict whether an edge of type $q$ exists between entities $x, y$ (\cref{fig:illustration}a), and the common methods are to calculate or learn a pair representation $\mathbf{h}^q(x, y)$ for prediction, \eg, using paths between two nodes or embedding methods or \acp{gnn}, while we explore using \acp{snn}. Many link prediction tasks are transductive, \ie, predicting new links on the training graph, and there is also the inductive setting where training and testing graphs have different entities but the same relation types.

\subsection{Synaptic Delay for Traditional Graph Algorithms}

\begin{figure*}[t!]
    \centering
    \includegraphics[width=0.68\linewidth]{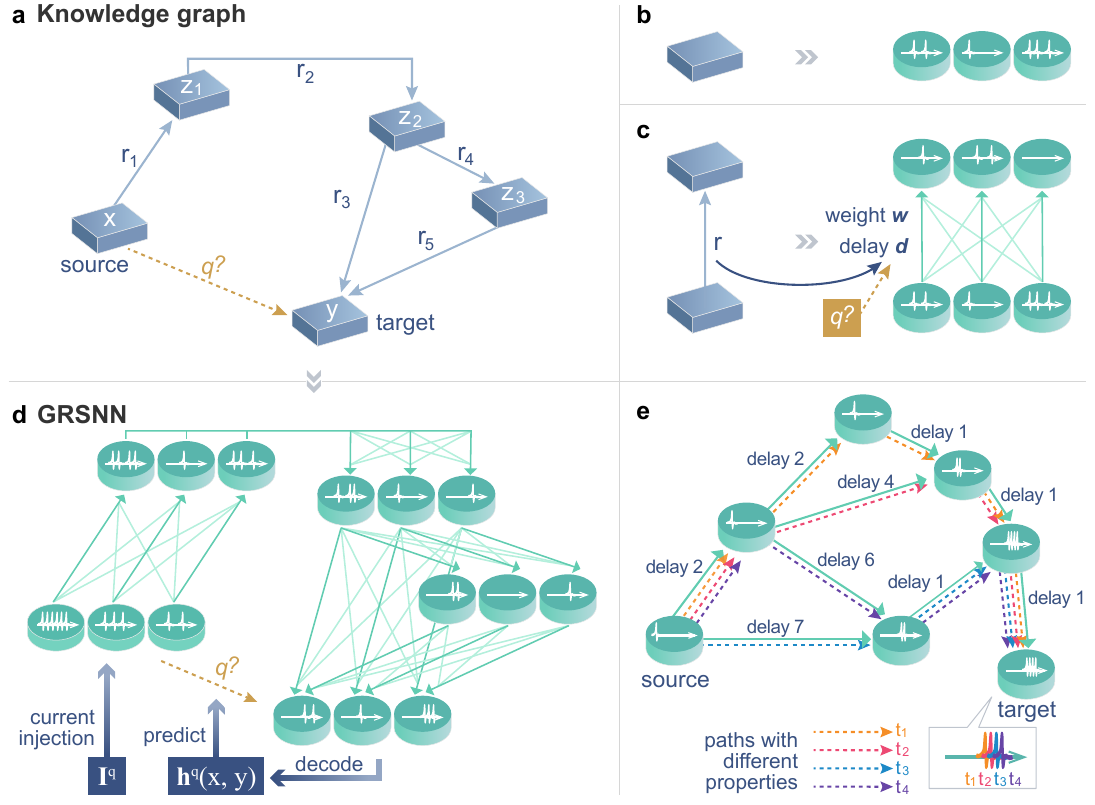}
    \vspace{-1mm}
    \caption{\textbf{Schematic of GRSNN.} (a) Illustration of the graph link prediction task. In GRSNN, (b) each graph entity node is associated with a cluster of spiking neurons, and (c) each relational edge corresponds to the synaptic connections between spiking neurons, with synaptic weight and delay. The weight can exhibit positive or negative values. The delay is contingent on the edge relation and query relation, representing the edge's property and the neuromodulation from the task goal. (d) Visualization of GRSNN. To predict a link, a constant current, dependent on the query relation, is injected into the spiking neurons of the source node, initiating the propagation of spike trains. After a specific time interval, the spike trains emanating from the target node are decoded to predict the query relation. (e) Depiction of the temporal domain serving as an additional dimension to process the properties of edges and paths in a network with more propagation paths. In the demonstrated network under a simplified setting where each input spike triggers an output spike for neurons, a spike from the source neuron will lead to four spikes from the target neuron, whose time varies corresponding to four propagation paths with different integrated properties of edges represented by synaptic delay.}
    \label{fig:illustration}
    \vspace{-2mm}
\end{figure*}

Some previous works show that the synaptic delay of \acp{snn} can be leveraged to solve traditional graph tasks, providing a parallelizable and efficient neuromorphic computing solution to graph algorithms~\citep{aimone2021provable}. 
For the traditional graph single-source shortest path problem, by assigning a neuron to each graph node and configuring the delay between neurons as the graph edge weight, \acp{snn} can parallelly simulate Dijkstra's algorithm. An example is shown in \cref{fig:illustration}e if we decode the spike train of the target neuron by the time to first spike. We will generalize the thought---delays in \acp{snn} can represent the properties of graph edges---to graph \ac{ai} reasoning tasks with neural generalization and advanced temporal coding with multiple temporal spikes for diverse paths.

\section{Graph Reasoning Spiking Neural Network}

In this section, we introduce our graph reasoning spiking neural networks. We first introduce the overview of our model in \cref{sec:model_overview}. Then in \cref{sec:snn_path_formulation}, we demonstrate that \ac{grsnn} can be viewed as a generalized path formulation for graph reasoning. In \cref{sec:comparison_gnn}, we discuss the comparison with graph neural networks. Finally, we introduce implementation details in \cref{sec:implementation_details}.

\subsection{Model Overview}\label{sec:model_overview}

The outline of \ac{grsnn} is depicted in \cref{fig:illustration}. Each graph node is assigned $n$ spiking neurons, representing each entity by a neuron population (\cref{fig:illustration}b). Synaptic connections, corresponding to relation links between entities, are characterized by weight and delay between neuron groups (\cref{fig:illustration}c). These synaptic properties, such as delay, are dependent on the graph edge relation and modulated by the query relation (task goal), allowing the integrated properties of paths to be reflected by the spiking time considering delays (\cref{fig:illustration}e). Unlike \acp{snn} for traditional graph tasks, we generalize the model to allow both positive and negative synaptic weights, acting as complementary transformations to learnable synaptic delays that are viewed as an additional dimension to process graph edges and paths.

For the link prediction task (\cref{fig:illustration}d), a constant current $\mathbf{I}^q$ is injected to the spiking neurons of the source node $x$ for a given query $q$ between nodes $x$ and $y$, generating spike trains. The network then propagates these spikes, and a spike train $\mathbf{s}_y^q(t)$ from the target node $y$’s neurons is obtained after a time interval. A decoding function $D$ calculates the pair representation $\mathbf{h}^q(x, y)=D(\mathbf{s}_y^q(t))$ for link prediction, and we primarily utilize temporal coding $D(\mathbf{s}_y^q(t))=\left(\sum_{\tau}\lambda^{\tau}\mathbf{s}_y^q[\tau]\right) / \left(\sum_{\tau}\lambda^\tau\right)$, emphasizing early spiking time. This corresponds to the decoding for various path formulations (refer to \cref{supp:sec:proof} for more details).

\subsection{GRSNN as Generalized Path Formulation}\label{sec:snn_path_formulation}

\ac{grsnn} serves as a neural generalization of the path formulation for graphs, allowing for the simultaneous consideration of all paths from a source node without the separate calculation of each one. Path formulation is important to graph reasoning due to better interpretability and inductive generalization ability~\citep{zhu2021neural,yang2017differentiable,sadeghian2019drum}. Traditional path-based algorithms calculate the pair representation between nodes $x$ and $y$ by considering paths from $x$ to $y$, formulated as a generalized accumulation of path representations~\citep{zhu2021neural}:
\vspace{-2mm}
\begin{equation}
    \small
    \mathbf{h}^q(x, y) = \bigoplus_{P\in\mathcal{P}_{xy}} \left(\bigotimes_{i=1}^{\vert P\vert}\mathbf{v}^q(e_i)\right),
\vspace{-1mm}
\end{equation}
where $\mathcal{P}_{xy}$ is the set of paths from $x$ to $y$, $e_i$ is the $i$-th edge on a path $P$, and $\mathbf{v}^q(e_i)$ is the edge representation (\eg, the transition probability of this edge). Various methods like Katz Index~\citep{katz1953new}, Personalized PageRank~\citep{page1998pagerank}, and Graph Distance~\citep{liben2007link} follow this modeling.

In \ac{grsnn}, spike trains propagate over time, with spikes at different times simultaneously maintaining all paths from the source node. The spike train of $y$ is:
\begin{equation}
\small
\begin{aligned}
    \mathbf{s}_y^q(t)&=f\left(\left\{\mathbf{s}_z^q(t), \mathbf{w}_{z, y}^q, \mathbf{d}_{z, y}^q \vert z\in \mathcal{N}(y)\right\}\right) \\
    &= \cdots = \overline{f}\left(\left\{ \mathbf{s}_x^q(t), \{\mathbf{w}_{e_i}^q, \mathbf{d}_{e_i}^q\}_{i=1}^{\vert P\vert} \vert P\in\mathcal{P}_{xy} \right\}\right), 
\end{aligned}
\end{equation}
where $f$ is the function of spiking neurons, $\mathbf{w}_{z,y}^q$ and $\mathbf{d}_{z,y}^q$ are synaptic weights and delays between nodes $z$ and $y$ given the query relation $q$, $\mathcal{N}(y)$ denotes the set of neighbors of $y$, and $\overline{f}$ denotes the general composite function for all paths. In some degenerated conditions, the time of a spike is the sum of edge delays on one path, allowing a decoding function $F$ to perform a general summation over all paths represented in the spike train. We show that, with specific settings, \ac{grsnn} can solve traditional path-based methods.

\begin{proposition}
    Katz Index, Personalized PageRank, and Graph Distance can be solved by GRSNN under specific settings.
\end{proposition}

The proof is detailed in \cref{supp:sec:proof}, focusing on the construction of appropriate delay and decoding functions. This proposition illustrates that \ac{grsnn} can degenerate to emulate traditional path-based algorithms. By employing parameterized synaptic delays for learnable edge representations, and additional parameters like synaptic weights for transformations in another dimension, \ac{grsnn} emerges as a neural generalization of the path formulation for graph reasoning. This sheds light on the capability of \acp{snn} to execute neuro-symbolic computation on graph paths utilizing spiking time and synaptic delay. Furthermore, \ac{grsnn}, as a generalization of path formulation, extends its important applicability to inductive settings and reasoning path interpretations, distinguishing it from entity embedding methods.

\subsection{Comparison with Graph Neural Networks}\label{sec:comparison_gnn}

The introduced \ac{grsnn} bears a resemblance to the widely-used message-passing \acp{gnn} in machine learning, both propagating messages between interconnected nodes. However, notable distinctions exist. 

First, \ac{grsnn} incorporates varied temporal synaptic delays in message passing, allowing for the encoding of relational information in spiking times with enhanced spatiotemporal processing. In contrast, \acp{gnn} uniformly propagate messages across all edges in each iteration.
Second, \ac{grsnn} disseminates temporal spike trains throughout the network, as opposed to \ac{gnn}'s real-valued activations. This not only facilitates the representation of multiple paths through diverse spiking times within a spike train but also promotes event-driven energy-efficient computation suitable for neuromorphic hardware.
Moreover, while \citet{zhu2021neural} interprets \ac{gnn} as a neural counterpart to the Bellman-Ford algorithm, \ac{grsnn} is perceived as a neural generalization of Dijkstra's algorithm. This parallel between artificial and brain-inspired neural networks in generalizing distinct classical algorithms for analogous objectives is intriguing.

Once the inherent differences are accounted for, \ac{grsnn} can also have a formulation analogous to \acp{gnn}. 
Specifically, at each discrete time step, every node (with spiking neurons) aggregates messages from neighbors. Assuming the sharing of synaptic weights across all edges, akin to \acp{gnn}, messages are represented by delayed spikes. The aggregation function then becomes a synthesis of the summation of all messages, a linear transformation, and the spike generation with neuronal dynamics of spiking neurons. Thus, for every node $z$, the following holds:
\begin{equation}
    \small
    \left\{
    \begin{aligned}
        &\mathbf{I}^q_z[t + 1] = e^{-\frac{\Delta \tau}{\tau_m}} \mathbf{I}^q_z[t] + \alpha \left(\mathbf{W} \sum_{k\in\mathcal{N}(z)}  \mathbf{s}^q_k[t - \mathbf{d}^q_{r}] + \mathbf{b}\right),\\
        &\mathbf{u}^q_z[t + 1] = e^{-\frac{\Delta \tau}{\tau_m}} \mathbf{u}^q_z[t] (1 - \mathbf{s}^q_z[t]) + \mathbf{I}^q_z[t + 1] + \mathbbm{1}_{z=x}\mathbf{I}^q,\\
        &\mathbf{s}^q_z[t + 1] = H(\mathbf{u}^q_z[t + 1] - V_{th}).\\
    \end{aligned}
    \right.
    \label{eq:grsnn}
\end{equation}
Here, $r$ signifies the relation from node $k$ to node $z$, $\mathbf{s}^q_k[t - \mathbf{d}^q_{r}]$ represents the vector of spikes with associated delays $\mathbf{d}^q_{r}$, and $\mathbbm{1}_{z=x}$ is an indicator for the current injection to the source node. The time steps can be viewed as the layers of \acp{gnn}, with shared weights and delays for all time steps. Consequently, the inference time and space complexity of \ac{grsnn} align closely with those of \acp{gnn}, except that they are proportional to the number of discrete time steps instead of \ac{gnn}'s layer number.

\subsection{Implementation Details}\label{sec:implementation_details}

\paragraph{Model Detail} In practice, our models predominantly adhere to \cref{eq:grsnn}. The set of learnable parameters encompasses $\mathbf{W}$ and $\mathbf{b}$, symbolizing a shared linear transformation of synaptic weights, and $\mathbf{d}^q_{r}$, denoting the delay between the spiking neurons of two nodes, contingent on their relation $r$ and the query relation $q$. Additionally, $\mathbf{r}$ signifies the embedding of relations, utilized for both current injection ($\mathbf{I}^q=\mathbf{r}^q$) and the ultimate link prediction with a parameterized function to predict links based on $\mathbf{h}^q(x, y)$ and $\mathbf{r}^q$. 
To differentiate the varying contributions of a relation (edge) in forecasting different query relations, we align with previous studies~\citep{zhu2021neural} to parameterize the edge representation of relation $r$ as a linear function over the query relation. This is then processed through a sigmoid function with a bound scale $\beta$ to serve as positive delays, \ie, $\mathbf{d}^q_{r}=\beta\sigma(\mathbf{W}_{r}\mathbf{r}^q+\mathbf{b}_{r})$. In the context of homogeneous graphs characterized by a singular relation, this simplifies to $\mathbf{d}^q_{r}=\beta\sigma(\mathbf{b}_{r})$. It undergoes quantization and is trained by the straight-through-estimator. Post-learning, it can be archived in a look-up table, obviating the need for nonlinear computations. This can be analogous to neuromodulation with a superior signal delineating the task objective. 

\begin{figure*}[t]
    \centering
    \includegraphics[width=\linewidth]{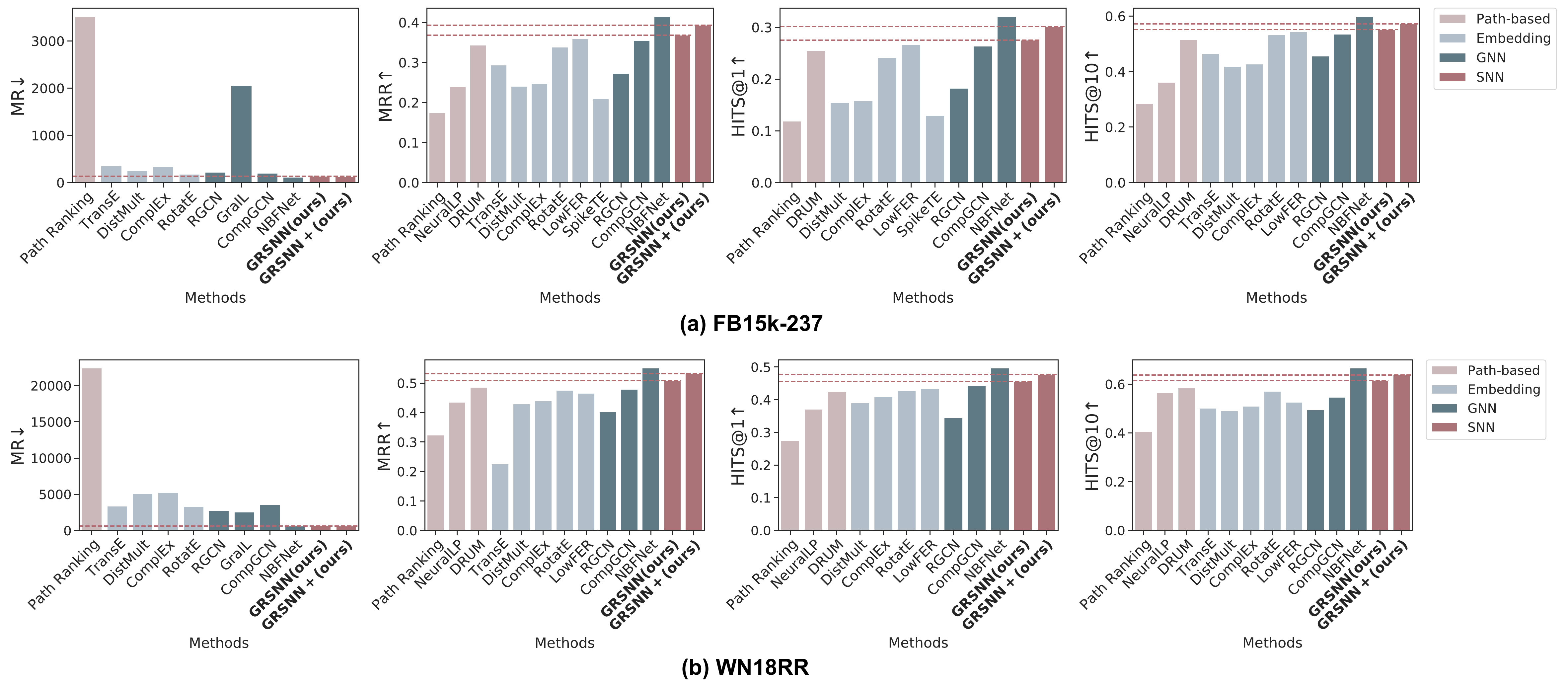}
    \vspace{-6mm}
    \caption{\textbf{Results of Transductive Knowledge Graph Completion on FB15k-237 and WN18RR.} Lower values are preferable for MR, while higher values are desirable for MRR, HITS@1, and HITS@10. Detailed values can be found in \cref{supp:sec:detailed_values}.} 
    \label{fig:transductive}
    \vspace{-2mm}
\end{figure*}

\paragraph{Link Prediction Detail} In line with prevalent practices for link prediction, the objective is to ascertain the likelihood of a triplet $(x, q, y)$, consisting of the source node, query relation, and target node. 
The procedure of our model to deduce a triplet $(x, q, y)$ commences with the propagation of spike trains across the graph to secure the pair representation $\mathbf{h}^q(x, y)$, and subsequently, the likelihood score is computed by a parameterized function $g$ given $\mathbf{h}^q(x, y)$, consistent with prior studies~\citep{zhu2021neural}. More details can be found in \cref{supp:sec:implementation_details}. The overarching procedure aligns with the conventional graph reasoning paradigm, with our primary focus being on the pivotal step of acquiring the pair representation through \ac{snn} propagation.

Regarding the training procedure, we adhere to the methodologies of preceding works~\citep{bordes2013translating,sunrotate,zhu2021neural}, generating negative samples by corrupting one entity in a positive triplet. Please refer to \cref{supp:sec:implementation_details} for more details. 

\section{Experiments}\label{sec:experiments}

In this section, we conduct experiments on transductive knowledge graph completion, inductive knowledge graph relation prediction, and homogeneous graph link prediction to evaluate the proposed GRSNN model. For knowledge graphs, we consider the commonly used FB15k-237~\citep{toutanova2015observed} and WN18RR~\citep{dettmers2018convolutional} with the standard transductive splits and inductive splits~\citep{teru2020inductive}. For homogeneous graphs, we consider Cora, Citeseer, and PubMed~\citep{sen2008collective}. The train/valid/test ratio of edges is 85:5:10 following the common practice, and the statistics of datasets can be found in \cref{supp:sec:dataset_statistics}. 

For evaluation of knowledge graph completion, we adhere to the filtered ranking protocol~\citep{bordes2013translating}, ranking a test triplet $(x, q, y)$ against all unseen negative triplets and report \ac{mr}, \ac{mrr}, and HITS@N. For inductive knowledge graph relation prediction, the evaluation adheres to the protocols outlined in the literature~\citep{teru2020inductive}, where 50 negative triplets are drawn for each positive one using the filtered ranking, and the results are reported as HITS@10. For homogeneous graph link prediction, we follow~\citet{kipf2016variational,zhu2021neural} to compare the positive edges against the same number of negative edges, and the results are quantified using \ac{auroc} and \ac{ap}.

More experimental details can be found in \cref{supp:sec:experimental_details}.

\subsection{Transductive Knowledge Graph Completion}
We initiate our evaluation with experiments on transductive knowledge graph completion to assess the efficacy of \ac{grsnn}. This task, illustrated in \cref{supp:sec:task_details}, involves predicting unseen relations between two existing entities in a knowledge graph and serves as a standard for assessing graph reasoning link prediction.  

\begin{table} [t]
	\centering
	\small
	\tabcolsep=1.5mm
	\caption{Results of knowledge graph completion on FB15k-237 by SNNs with different methods to represent relation information. For MR, the lower the better. For MRR, HITS@1, HITS@3, and HITS@10, the higher the better. }
	\begin{tabular}{c|ccccc}
		\toprule[1pt]
        Method & MR$\downarrow$ & MRR$\uparrow$ & H@1$\uparrow$ & H@3$\uparrow$ & H@10$\uparrow$\\
		\midrule[0.5pt]
		None & 396 & 0.204 & 0.119 & 0.226 & 0.380\\
        Synaptic weight & 197 & 0.311 & 0.220 & 0.343 & 0.491\\
        Synaptic delay & \textbf{139} & \textbf{0.368} & \textbf{0.275} & \textbf{0.407} & \textbf{0.551}\\
		\bottomrule[1pt]
	\end{tabular}
	\label{table:synaptic_delay}
\end{table}

\begin{figure*}[ht]
    \centering
    \includegraphics[width=0.85\linewidth]{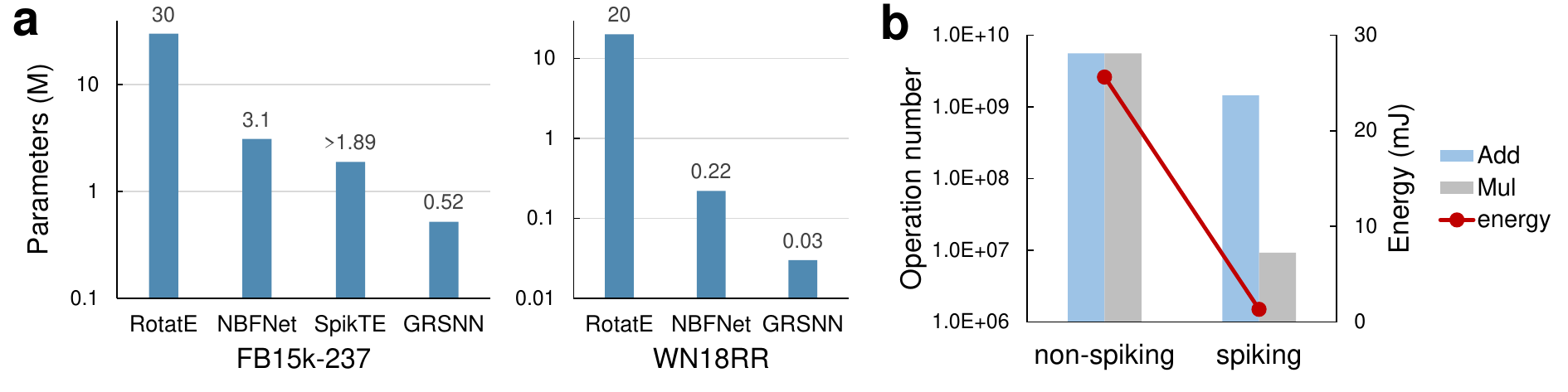}
    \vspace{-2mm}
    \caption{\textbf{Analytical Results for GRSNN.} (a) Log-scale comparison of the parameter quantities across different methods, demonstrating the enhanced parameter efficiency of GRSNN. (b) Theoretical estimations of the number of ADD and MUL operations (log scale) and energy consumption on FB15k-237. GRSNN can achieve approximately $20\times$ energy reduction compared to its non-spiking counterpart.} 
    \label{fig:analysis_results}
    \vspace{-2mm}
\end{figure*}

\paragraph*{Advantage of synaptic delay} 

We investigate the role of synaptic delay in encoding relational information for reasoning, illustrated in \cref{table:synaptic_delay}. Our experiments contrast two baselines. The first baseline does not encode edge relations, focusing solely on the existence of edges. The second encodes edge relations with an additional relation-dependent term in synaptic weights, eschewing synaptic delay, reminiscent of the DistMult message function in \ac{gnn}. More details are provided in the \cref{supp:sec:experimental_details}. The results, presented in \cref{table:synaptic_delay}, highlight that synaptic delay significantly excels over the baselines, accentuating the merits of incorporating temporal processing with delays in bio-inspired models for effective relational reasoning.

\paragraph*{Comparison with prevalent machine learning methods} 

We juxtapose the performance of our bio-inspired \ac{grsnn} with various machine learning methods, including path-based, embedding, and \ac{gnn} methods, as depicted in \cref{fig:transductive}, to underscore its efficacy in knowledge graph reasoning. We derive the results of preceding methods~\citep{zhu2021neural,vashishthcomposition,dold2022relational}. In essence, \ac{grsnn} secures competitive results, surpassing the majority of machine learning methods across all metrics, thereby attesting to the effectiveness of bio-inspired models in solving human-like advanced knowledge reasoning tasks. NBFNet attains superior performance by employing numerous \ac{gnn} tricks that we deliberately omitted to preserve the inherent properties of \acp{snn}. If we further integrate some techniques (refer to \cref{supp:sec:experimental_details}), our model, denoted as \textit{GRSNN+} in \cref{fig:transductive}, also achieves a better performance. Note that the proposed \ac{grsnn} prioritizes bio-plausibility, delivering promising performance with augmented efficiency, as will be analyzed in the following.

\subsection{Analysis Results}

\paragraph{Parameter amount} \cref{fig:analysis_results}a contrasts the parameter quantities of several representative methods, highlighting the notable parameter efficiency of \ac{grsnn} in achieving competitive performance compared to other methods.

\paragraph{Theoretical estimation of energy} \ac{grsnn} leverages the energy efficiency inherent to \acp{snn} through spike-based computation. On the test set of FB15k-237, the model exhibits a spike rate---the average spike count per discrete time step—of approximately 0.258. This translates to roughly a $4\times$ reduction in synaptic operations compared to equivalent real-valued neural networks. Given that spikes necessitate only \ac{ac} operations as opposed to \ac{mac} operations, there is a substantial reduction in energy costs, as evidenced by the energy consumption of 32-bit FP \ac{mac} and \ac{ac} operations on a 45 nm CMOS processor being 4.6 pJ and 0.9 pJ, respectively~\citep{horowitz20141}. \cref{fig:analysis_results}b provides a concise theoretical estimation of the number of addition and multiplication operations and the associated energy requirements, with the multiplication in \acp{snn} arising due to leaky neuronal dynamics (please refer to \cref{supp:sec:experimental_details} for calculation details). Based on these estimations, a potential $20\times$ energy reduction is foreseeable, and under certain conditions where \ac{ac} can be $31\times$ cheaper than \ac{mac}~\citep{yin2021accuratenmi,horowitz20141}, this could extend to around $100\times$. 

Note that there can also be costs from synaptic delay. We consider the Ring Buffer as a potential implementation, which is commonly used by digital neuromorphic platforms and analyzed~\citep{patino2023empirical}. The additional energy overhead will account for an extremely small proportion of energy---it is estimated as 0.004 mJ, while the energy for synaptic operations estimated above is 1.337 mJ (please refer to \cref{supp:sec:experimental_details} for calculation details), and this conclusion is consistent with \citet{patino2023empirical}. Therefore, the costs of synaptic delay do not affect the substantial energy efficiency.

More spike rate statistics on different datasets or tasks are presented in \cref{table:spike_rate}, showing that spikes are even sparser on other datasets or tasks introduced in the following. 
This underscores the substantial potential of \ac{grsnn} in enhancing energy efficiency by one to two orders of magnitude.

\begin{figure*}[ht]
    \centering
    \includegraphics[width=0.88\linewidth]{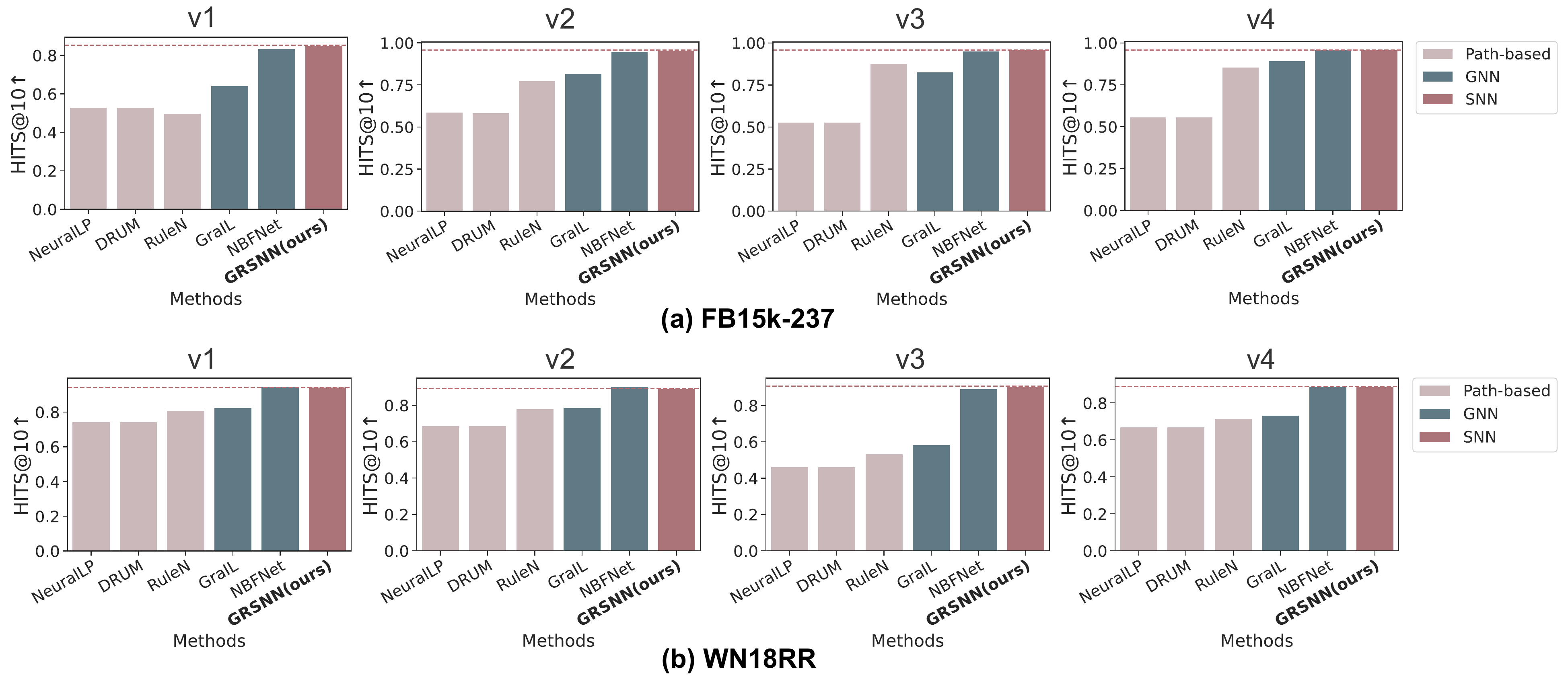}
    \vspace{-2mm}
    \caption{\textbf{Results of Inductive Relation Prediction on FB15k-237 and WN18RR.} v1-v4 correspond to the four standard versions of inductive splits. Detailed values can be found in \cref{supp:sec:detailed_values}.} 
    \label{fig:inductive}
\end{figure*}

\begin{figure*}[ht]
    \centering
    \includegraphics[width=0.95\linewidth]{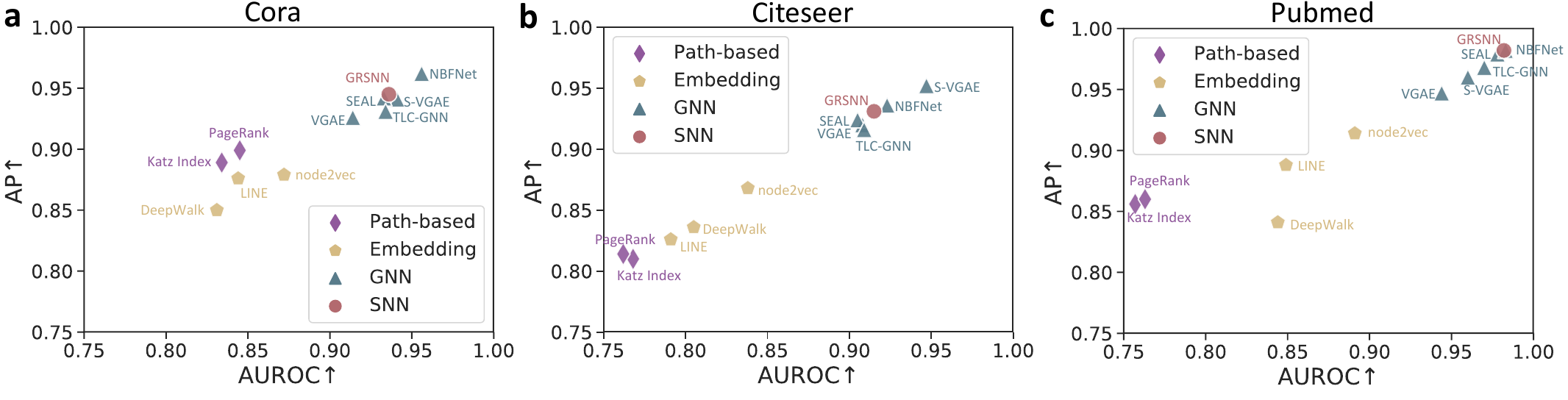}
    \vspace{-4mm}
    \caption{\textbf{Results of Homogeneous Graph Link Prediction on Cora, Citeseer, and PubMed.}  Detailed values are in \cref{supp:sec:detailed_values}.}
    \label{fig:homogeneous}
    \vspace{-2mm}
\end{figure*}

\begin{table*}[ht]
    \centering
    \small
    \caption{\textbf{More spike rate statistics on different datasets.}}
    \label{table:spike_rate}
    \resizebox{\linewidth}{!}{%
    \begin{tabular}{cc|cccccccc|ccc}
         \toprule
         \multicolumn{2}{c|}{Trans. Know. Graph Compl.} & \multicolumn{8}{c|}{Induc. Relat. Pred.} & \multicolumn{3}{c}{Homo. Graph Link Pred.}\\
         FB15k-237 & WN18RR & FB. v1 & FB. v2 & FB. v3 & FB. v4 & WN. v1 & WN. v2 & WN. v3 & WN. v4 & Cora & CiteSeer & PubMed \\
         \midrule
         0.258 & 0.191 & 0.176 & 0.189 & 0.257 & 0.245 & 0.175 & 0.165 & 0.172 & 0.149 & 0.082 & 0.074 & 0.189\\
         \bottomrule
    \end{tabular}
    }
\end{table*}

\paragraph{Interpretability} To demonstrate the interpretability of \ac{grsnn} as neural-generalized path formulation, in \cref{supp:sec:interpretability}, we visualize the reasoning paths for the final predictions of several examples, based on edge and path importance, determined by the gradient of the prediction \wrt edges, and beam search for paths of higher importance (refer to \cref{supp:sec:experimental_details} for details). Results show that GRSNN is adept at discerning relation relevances and exploiting transitions and analogs. 

More analysis results such as the impact of temporal discretization steps and the impact of neuron number and parameter amount are in \cref{supp:sec:more_results}.

\subsection{Inductive Relation Prediction}

Experiments are also conducted on inductive relation prediction to assess the efficacy of GRSNN. Unlike the transductive setting, which focuses on predicting new links within the training knowledge graph, inductive prediction strives to extrapolate the ability to predict relations from the training graph to a distinct testing graph. This testing graph encompasses different entities but retains the same relation types, as illustrated in \cref{supp:sec:task_details}, demonstrating the ability to generalize relational reasoning to new conditions. Traditional entity embedding methods falter under this condition, whereas \ac{grsnn}, being a generalized form of path formulation, adeptly manages it.

The outcomes, depicted in \cref{fig:inductive}, reveal that \ac{grsnn} surpasses the performance of most machine learning methods in inductive settings, underscoring its proficiency in generalizing reasoning to new entities. The spike rate statistics in \cref{table:spike_rate} also indicate the potential for energy efficiency.

\subsection{Homogeneous Graph Link Prediction}

We also assess the \ac{grsnn} in the context of link prediction tasks for standard homogeneous graphs, illustrating its versatility across diverse application domains. Homogeneous graphs are essentially a subset of knowledge graphs, characterized by a singular type of relation, \ie, the presence of graph edges, and are ubiquitously observed. In such instances, the representation of edges remains consistent across the graph, and the \ac{grsnn} primarily leverages the information pertaining to graph distance in spiking time, as opposed to relation-specific information.

The results shown in \cref{fig:homogeneous} reveal that \ac{grsnn} manifests competitive performance in comparison to other proficient machine learning models, underscoring its efficacy. The sparse spike rates presented in \cref{table:spike_rate} further highlight the efficiency potential.

\section{Discussion and Conclusion}

This study demonstrates the potential of bio-inspired \acp{snn} in addressing graph reasoning through the innovative use of synaptic delay and spiking time. We introduced \ac{grsnn}, a model that employs synaptic delays to encode relation information of graph edges and utilizes the temporal domain as an additional dimension for processing graph path properties. This approach can be perceived as a neural generalization of the path formulation with better inductive generalization ability and interpretability. It provides insights into the capabilities of networks with biological neuron models to efficiently facilitate neuro-symbolic reasoning in tasks central to human intelligence, such as relational reasoning of concepts. Additionally, it explores the enhanced role that spiking time can play in \ac{ai} applications. The promising performance and substantial theoretical energy efficiency of our model underscore the potential of \acp{snn} in a wider array of applications such as efficient reasoning.

Our approach to temporal coding of spike trains assigns varying weights to different times, which is similar to the methodology in \citet{stockl2021optimized} but in our model, earlier spikes are designed to receive higher weights, which also integrates concepts from the time to first spike paradigm~\citep{mostafa2017supervised}. Distinct to these works, our focus extends beyond individual neuron temporal coding to encompass the network level, allowing spiking time to integrate path properties during network propagation, and enabling multiple spikes to represent diverse paths globally. Unlike prior studies on traditional graph algorithms~\citep{aimone2021provable}, which primarily target the shortest path task, our work delves into the multifaceted realm of graph \ac{ai} tasks with multiple temporal spikes for diverse paths. Together, our work offers a fresh perspective on temporal information processing in \acp{snn}.

This study marks an initial exploration of utilizing \acp{snn} for graph reasoning, leveraging the temporal domain, and opens avenues for numerous future directions. 
First, the reliance on discrete simulation and backpropagation through time for training \acp{snn} is resource-intensive, especially for long simulation times with small discrete intervals. Therefore, this work primarily considers $T=10$ discrete time steps, which could sacrifice more precise temporal information. Enhancements in simulation or training methodologies at hardware/coding/algorithm levels may yield improved results given more accurate temporal information. Additionally, the exploration of online training methods conducive to on-chip learning of \acp{snn}~\citep{bellec2020solution,xiao2022online} for learning synaptic delays may offer insights into efficient and more biologically plausible learning of our model. 
Second, to fulfill the properties of \acp{snn}, many advanced \ac{gnn} strategies may not be incorporated, such as intricate message and aggregate functions and elaborate network structures. The investigation of \ac{snn}-compatible strategies may potentially bridge the performance gap, \eg, heterogeneous neurons and different neuron dynamics may provide more powerful computational properties~\cite{chakraborty2023heterogeneous,bellec2018long}. 
Last, given the prevalence of graph tasks in \ac{ai} applications, future studies could delve into wider applications of graph reasoning such as drug or material design, investigating the potential of bio-inspired models for efficient applications.

In conclusion, our study illustrates the capability of brain-inspired \acp{snn} in efficient symbolic graph reasoning, emphasizing the enhanced role of the temporal domain. Given their neuromorphic attributes, \acp{snn} are poised to achieve substantial energy efficiency and high parallelism on spike-based neuromorphic hardware. It is our aspiration that this research serves as a catalyst for deeper insights and wider applications of biologically inspired efficient \acp{snn}.

\section*{Acknowledgements}

Z. Lin was supported by National Key R\&D Program of China (2022ZD0160300), the NSF China (No. 62276004), and Qualcomm. D. He was supported by National Science Foundation of China (NSFC62376007).

\section*{Impact Statement}

This paper presents work whose goal is to advance the field of 
Machine Learning. There are many potential societal consequences 
of our work, none which we feel must be specifically highlighted here.


\bibliography{reference_header,GRSNN}

\begin{thebibliography}{68}
\providecommand{\natexlab}[1]{#1}
\providecommand{\url}[1]{\texttt{#1}}
\expandafter\ifx\csname urlstyle\endcsname\relax
  \providecommand{\doi}[1]{doi: #1}\else
  \providecommand{\doi}{doi: \begingroup \urlstyle{rm}\Url}\fi

\bibitem[Aimone et~al.(2021)Aimone, Ho, Parekh, Phillips, Pinar, Severa, and
  Wang]{aimone2021provable}
Aimone, J.~B., Ho, Y., Parekh, O., Phillips, C.~A., Pinar, A., Severa, W., and
  Wang, Y.
\newblock Provable advantages for graph algorithms in spiking neural networks.
\newblock In \emph{ACM Symposium on Parallelism in Algorithms and Architectures
  (SPAA)}, 2021.

\bibitem[Amin et~al.(2020)Amin, Varanasi, Dunfield, and
  Neumann]{amin2020lowfer}
Amin, S., Varanasi, S., Dunfield, K.~A., and Neumann, G.
\newblock Lowfer: Low-rank bilinear pooling for link prediction.
\newblock In \emph{International Conference on Machine Learning (ICML)}, 2020.

\bibitem[Bellec et~al.(2018)Bellec, Salaj, Subramoney, Legenstein, and
  Maass]{bellec2018long}
Bellec, G., Salaj, D., Subramoney, A., Legenstein, R., and Maass, W.
\newblock Long short-term memory and learning-to-learn in networks of spiking
  neurons.
\newblock In \emph{Advances in Neural Information Processing Systems
  (NeurIPS)}, 2018.

\bibitem[Bellec et~al.(2020)Bellec, Scherr, Subramoney, Hajek, Salaj,
  Legenstein, and Maass]{bellec2020solution}
Bellec, G., Scherr, F., Subramoney, A., Hajek, E., Salaj, D., Legenstein, R.,
  and Maass, W.
\newblock A solution to the learning dilemma for recurrent networks of spiking
  neurons.
\newblock \emph{Nature Communications}, 11\penalty0 (1):\penalty0 1--15, 2020.

\bibitem[Bordes et~al.(2013)Bordes, Usunier, Garcia-Duran, Weston, and
  Yakhnenko]{bordes2013translating}
Bordes, A., Usunier, N., Garcia-Duran, A., Weston, J., and Yakhnenko, O.
\newblock Translating embeddings for modeling multi-relational data.
\newblock In \emph{Advances in Neural Information Processing Systems
  (NeurIPS)}, 2013.

\bibitem[Chakraborty \& Mukhopadhyay(2023)Chakraborty and
  Mukhopadhyay]{chakraborty2023heterogeneous}
Chakraborty, B. and Mukhopadhyay, S.
\newblock Heterogeneous neuronal and synaptic dynamics for spike-efficient
  unsupervised learning: Theory and design principles.
\newblock In \emph{International Conference on Learning Representations
  (ICLR)}, 2023.

\bibitem[Comsa et~al.(2020)Comsa, Potempa, Versari, Fischbacher, Gesmundo, and
  Alakuijala]{comsa2020temporal}
Comsa, I.~M., Potempa, K., Versari, L., Fischbacher, T., Gesmundo, A., and
  Alakuijala, J.
\newblock Temporal coding in spiking neural networks with alpha synaptic
  function.
\newblock In \emph{IEEE International Conference on Acoustics, Speech and
  Signal Processing (ICASSP)}, 2020.

\bibitem[Davidson et~al.(2018)Davidson, Falorsi, De~Cao, Kipf, and
  Tomczak]{davidson2018hyperspherical}
Davidson, T.~R., Falorsi, L., De~Cao, N., Kipf, T., and Tomczak, J.~M.
\newblock Hyperspherical variational auto-encoders.
\newblock In \emph{Conference on Uncertainty in Artificial Intelligence (UAI)},
  2018.

\bibitem[Davies et~al.(2018)Davies, Srinivasa, Lin, Chinya, Cao, Choday, Dimou,
  Joshi, Imam, Jain, et~al.]{davies2018loihi}
Davies, M., Srinivasa, N., Lin, T.-H., Chinya, G., Cao, Y., Choday, S.~H.,
  Dimou, G., Joshi, P., Imam, N., Jain, S., et~al.
\newblock Loihi: A neuromorphic manycore processor with on-chip learning.
\newblock \emph{IEEE Micro}, 38\penalty0 (1):\penalty0 82--99, 2018.

\bibitem[Dettmers et~al.(2018)Dettmers, Minervini, Stenetorp, and
  Riedel]{dettmers2018convolutional}
Dettmers, T., Minervini, P., Stenetorp, P., and Riedel, S.
\newblock Convolutional 2d knowledge graph embeddings.
\newblock In \emph{AAAI Conference on Artificial Intelligence (AAAI)}, 2018.

\bibitem[Dold(2022)]{dold2022relational}
Dold, D.
\newblock Relational representation learning with spike trains.
\newblock In \emph{International Joint Conference on Neural Networks (IJCNN)},
  2022.

\bibitem[Dold \& Garrido(2021)Dold and Garrido]{dold2021spike}
Dold, D. and Garrido, J.~S.
\newblock Spike: Spike-based embeddings for multi-relational graph data.
\newblock In \emph{International Joint Conference on Neural Networks (IJCNN)},
  2021.

\bibitem[Fang et~al.(2022)Fang, Zeng, Tang, Wang, Liang, and
  Liu]{fang2022brain}
Fang, H., Zeng, Y., Tang, J., Wang, Y., Liang, Y., and Liu, X.
\newblock Brain-inspired graph spiking neural networks for commonsense
  knowledge representation and reasoning.
\newblock \emph{arXiv preprint arXiv:2207.05561}, 2022.

\bibitem[Fang et~al.(2021)Fang, Yu, Chen, Masquelier, Huang, and
  Tian]{Fang_2021_ICCV}
Fang, W., Yu, Z., Chen, Y., Masquelier, T., Huang, T., and Tian, Y.
\newblock Incorporating learnable membrane time constant to enhance learning of
  spiking neural networks.
\newblock In \emph{International Conference on Computer Vision (ICCV)}, 2021.

\bibitem[Grover \& Leskovec(2016)Grover and Leskovec]{grover2016node2vec}
Grover, A. and Leskovec, J.
\newblock node2vec: Scalable feature learning for networks.
\newblock In \emph{ACM SIGKDD International Conference on Knowledge Discovery
  and Data Mining (KDD)}, 2016.

\bibitem[Hammouamri et~al.(2024)Hammouamri, Khalfaoui-Hassani, and
  Masquelier]{hammouamri2024learning}
Hammouamri, I., Khalfaoui-Hassani, I., and Masquelier, T.
\newblock Learning delays in spiking neural networks using dilated convolutions
  with learnable spacings.
\newblock In \emph{International Conference on Learning Representations
  (ICLR)}, 2024.

\bibitem[Horowitz(2014)]{horowitz20141}
Horowitz, M.
\newblock 1.1 computing's energy problem (and what we can do about it).
\newblock In \emph{IEEE International Solid-State Circuits Conference (ISSCC)},
  2014.

\bibitem[Huxter et~al.(2003)Huxter, Burgess, and
  O'Keefe]{huxter2003independent}
Huxter, J., Burgess, N., and O'Keefe, J.
\newblock Independent rate and temporal coding in hippocampal pyramidal cells.
\newblock \emph{Nature}, 425\penalty0 (6960):\penalty0 828--832, 2003.

\bibitem[Katz(1953)]{katz1953new}
Katz, L.
\newblock A new status index derived from sociometric analysis.
\newblock \emph{Psychometrika}, 18\penalty0 (1):\penalty0 39--43, 1953.

\bibitem[Kemp \& Tenenbaum(2008)Kemp and Tenenbaum]{kemp2008discovery}
Kemp, C. and Tenenbaum, J.~B.
\newblock The discovery of structural form.
\newblock \emph{Proceedings of the National Academy of Sciences (PNAS)},
  105\penalty0 (31):\penalty0 10687--10692, 2008.

\bibitem[Kipf \& Welling(2016)Kipf and Welling]{kipf2016variational}
Kipf, T.~N. and Welling, M.
\newblock Variational graph auto-encoders.
\newblock \emph{arXiv preprint arXiv:1611.07308}, 2016.

\bibitem[Lao \& Cohen(2010)Lao and Cohen]{lao2010relational}
Lao, N. and Cohen, W.~W.
\newblock Relational retrieval using a combination of path-constrained random
  walks.
\newblock \emph{Machine Learning}, 81:\penalty0 53--67, 2010.

\bibitem[Li et~al.(2023)Li, Yu, Zhu, Chen, Yu, Zheng, Tian, Wu, and
  Meng]{li2023scaling}
Li, J., Yu, Z., Zhu, Z., Chen, L., Yu, Q., Zheng, Z., Tian, S., Wu, R., and
  Meng, C.
\newblock Scaling up dynamic graph representation learning via spiking neural
  networks.
\newblock In \emph{AAAI Conference on Artificial Intelligence (AAAI)}, 2023.

\bibitem[Li et~al.(2021)Li, Deng, Dong, Gong, and Gu]{li2021free}
Li, Y., Deng, S., Dong, X., Gong, R., and Gu, S.
\newblock A free lunch from ann: Towards efficient, accurate spiking neural
  networks calibration.
\newblock In \emph{International Conference on Machine Learning (ICML)}, 2021.

\bibitem[Liben-Nowell \& Kleinberg(2007)Liben-Nowell and
  Kleinberg]{liben2007link}
Liben-Nowell, D. and Kleinberg, J.
\newblock The link-prediction problem for social networks.
\newblock \emph{Journal of the American Society for Information Science and
  Technology}, 58\penalty0 (7):\penalty0 1019--1031, 2007.

\bibitem[Lv et~al.(2023)Lv, Xu, and Zheng]{lv2023spiking}
Lv, C., Xu, J., and Zheng, X.
\newblock Spiking convolutional neural networks for text classification.
\newblock In \emph{International Conference on Learning Representations
  (ICLR)}, 2023.

\bibitem[Maass(1997)]{maass1997networks}
Maass, W.
\newblock Networks of spiking neurons: the third generation of neural network
  models.
\newblock \emph{Neural Networks}, 10\penalty0 (9):\penalty0 1659--1671, 1997.

\bibitem[Meilicke et~al.(2018)Meilicke, Fink, Wang, Ruffinelli, Gemulla, and
  Stuckenschmidt]{meilicke2018fine}
Meilicke, C., Fink, M., Wang, Y., Ruffinelli, D., Gemulla, R., and
  Stuckenschmidt, H.
\newblock Fine-grained evaluation of rule-and embedding-based systems for
  knowledge graph completion.
\newblock In \emph{International Semantic Web Conference (ISWC)}, 2018.

\bibitem[Meng et~al.(2022{\natexlab{a}})Meng, Xiao, Yan, Wang, Lin, and
  Luo]{meng2022training}
Meng, Q., Xiao, M., Yan, S., Wang, Y., Lin, Z., and Luo, Z.-Q.
\newblock Training high-performance low-latency spiking neural networks by
  differentiation on spike representation.
\newblock In \emph{Conference on Computer Vision and Pattern Recognition
  (CVPR)}, 2022{\natexlab{a}}.

\bibitem[Meng et~al.(2022{\natexlab{b}})Meng, Yan, Xiao, Wang, Lin, and
  Luo]{meng2022trainingnn}
Meng, Q., Yan, S., Xiao, M., Wang, Y., Lin, Z., and Luo, Z.-Q.
\newblock Training much deeper spiking neural networks with a small number of
  time-steps.
\newblock \emph{Neural Networks}, 153:\penalty0 254--268, 2022{\natexlab{b}}.

\bibitem[Mostafa(2017)]{mostafa2017supervised}
Mostafa, H.
\newblock Supervised learning based on temporal coding in spiking neural
  networks.
\newblock \emph{IEEE Transactions on Neural Networks and Learning Systems
  (TNNLS)}, 29\penalty0 (7):\penalty0 3227--3235, 2017.

\bibitem[Nickel et~al.(2015)Nickel, Murphy, Tresp, and
  Gabrilovich]{nickel2015review}
Nickel, M., Murphy, K., Tresp, V., and Gabrilovich, E.
\newblock A review of relational machine learning for knowledge graphs.
\newblock \emph{Proceedings of the IEEE}, 104\penalty0 (1):\penalty0 11--33,
  2015.

\bibitem[Page et~al.(1999)Page, Brin, Motwani, and Winograd]{page1998pagerank}
Page, L., Brin, S., Motwani, R., and Winograd, T.
\newblock The pagerank citation ranking: Bringing order to the web.
\newblock Technical report, Stanford InforLab, 1999.

\bibitem[Pati{\~n}o-Saucedo et~al.(2023)Pati{\~n}o-Saucedo, Yousefzadeh, Tang,
  Corradi, Linares-Barranco, and Sifalakis]{patino2023empirical}
Pati{\~n}o-Saucedo, A., Yousefzadeh, A., Tang, G., Corradi, F.,
  Linares-Barranco, B., and Sifalakis, M.
\newblock Empirical study on the efficiency of spiking neural networks with
  axonal delays, and algorithm-hardware benchmarking.
\newblock In \emph{IEEE International Symposium on Circuits and Systems
  (ISCAS)}. IEEE, 2023.

\bibitem[Pei et~al.(2019)Pei, Deng, Song, Zhao, Zhang, Wu, Wang, Zou, Wu, He,
  et~al.]{pei2019towards}
Pei, J., Deng, L., Song, S., Zhao, M., Zhang, Y., Wu, S., Wang, G., Zou, Z.,
  Wu, Z., He, W., et~al.
\newblock {Towards artificial general intelligence with hybrid Tianjic chip
  architecture}.
\newblock \emph{Nature}, 572\penalty0 (7767):\penalty0 106--111, 2019.

\bibitem[Perozzi et~al.(2014)Perozzi, Al-Rfou, and Skiena]{perozzi2014deepwalk}
Perozzi, B., Al-Rfou, R., and Skiena, S.
\newblock Deepwalk: Online learning of social representations.
\newblock In \emph{ACM SIGKDD International Conference on Knowledge Discovery
  and Data Mining (KDD)}, 2014.

\bibitem[Rao et~al.(2022)Rao, Plank, Wild, and Maass]{rao2022long}
Rao, A., Plank, P., Wild, A., and Maass, W.
\newblock A long short-term memory for {AI} applications in spike-based
  neuromorphic hardware.
\newblock \emph{Nature Machine Intelligence}, 4\penalty0 (5):\penalty0
  467--479, 2022.

\bibitem[Reinagel \& Reid(2000)Reinagel and Reid]{reinagel2000temporal}
Reinagel, P. and Reid, R.~C.
\newblock Temporal coding of visual information in the thalamus.
\newblock \emph{Journal of Neuroscience}, 20\penalty0 (14):\penalty0
  5392--5400, 2000.

\bibitem[Roy et~al.(2019)Roy, Jaiswal, and Panda]{roy2019towards}
Roy, K., Jaiswal, A., and Panda, P.
\newblock Towards spike-based machine intelligence with neuromorphic computing.
\newblock \emph{Nature}, 575\penalty0 (7784):\penalty0 607--617, 2019.

\bibitem[Rueckauer et~al.(2017)Rueckauer, Lungu, Hu, Pfeiffer, and
  Liu]{rueckauer2017conversion}
Rueckauer, B., Lungu, I.-A., Hu, Y., Pfeiffer, M., and Liu, S.-C.
\newblock Conversion of continuous-valued deep networks to efficient
  event-driven networks for image classification.
\newblock \emph{Frontiers in Neuroscience}, 11:\penalty0 682, 2017.

\bibitem[Sadeghian et~al.(2019)Sadeghian, Armandpour, Ding, and
  Wang]{sadeghian2019drum}
Sadeghian, A., Armandpour, M., Ding, P., and Wang, D.~Z.
\newblock Drum: End-to-end differentiable rule mining on knowledge graphs.
\newblock In \emph{Advances in Neural Information Processing Systems
  (NeurIPS)}, 2019.

\bibitem[Santoro et~al.(2017)Santoro, Raposo, Barrett, Malinowski, Pascanu,
  Battaglia, and Lillicrap]{santoro2017simple}
Santoro, A., Raposo, D., Barrett, D.~G., Malinowski, M., Pascanu, R.,
  Battaglia, P., and Lillicrap, T.
\newblock A simple neural network module for relational reasoning.
\newblock In \emph{Advances in Neural Information Processing Systems
  (NeurIPS)}, 2017.

\bibitem[Schlichtkrull et~al.(2018)Schlichtkrull, Kipf, Bloem, Van Den~Berg,
  Titov, and Welling]{schlichtkrull2018modeling}
Schlichtkrull, M., Kipf, T.~N., Bloem, P., Van Den~Berg, R., Titov, I., and
  Welling, M.
\newblock Modeling relational data with graph convolutional networks.
\newblock In \emph{European Semantic Web Conference (ESWC)}, 2018.

\bibitem[Schuman et~al.(2022)Schuman, Kulkarni, Parsa, Mitchell, Kay,
  et~al.]{schuman2022opportunities}
Schuman, C.~D., Kulkarni, S.~R., Parsa, M., Mitchell, J.~P., Kay, B., et~al.
\newblock Opportunities for neuromorphic computing algorithms and applications.
\newblock \emph{Nature Computational Science}, 2\penalty0 (1):\penalty0 10--19,
  2022.

\bibitem[Sen et~al.(2008)Sen, Namata, Bilgic, Getoor, Galligher, and
  Eliassi-Rad]{sen2008collective}
Sen, P., Namata, G., Bilgic, M., Getoor, L., Galligher, B., and Eliassi-Rad, T.
\newblock Collective classification in network data.
\newblock \emph{AI Magazine}, 29\penalty0 (3):\penalty0 93--93, 2008.

\bibitem[Shrestha \& Orchard(2018)Shrestha and Orchard]{shrestha2018slayer}
Shrestha, S.~B. and Orchard, G.
\newblock Slayer: spike layer error reassignment in time.
\newblock In \emph{Advances in Neural Information Processing Systems
  (NeurIPS)}, 2018.

\bibitem[St{\"o}ckl \& Maass(2021)St{\"o}ckl and Maass]{stockl2021optimized}
St{\"o}ckl, C. and Maass, W.
\newblock Optimized spiking neurons can classify images with high accuracy
  through temporal coding with two spikes.
\newblock \emph{Nature Machine Intelligence}, 3\penalty0 (3):\penalty0
  230--238, 2021.

\bibitem[Sun et~al.(2019)Sun, Deng, Nie, and Tang]{sunrotate}
Sun, Z., Deng, Z.-H., Nie, J.-Y., and Tang, J.
\newblock Rotate: Knowledge graph embedding by relational rotation in complex
  space.
\newblock In \emph{International Conference on Learning Representations
  (ICLR)}, 2019.

\bibitem[Tang et~al.(2015)Tang, Qu, Wang, Zhang, Yan, and Mei]{tang2015line}
Tang, J., Qu, M., Wang, M., Zhang, M., Yan, J., and Mei, Q.
\newblock Line: Large-scale information network embedding.
\newblock In \emph{International Conference on World Wide Web (WWW)}, 2015.

\bibitem[Teru et~al.(2020)Teru, Denis, and Hamilton]{teru2020inductive}
Teru, K., Denis, E., and Hamilton, W.
\newblock Inductive relation prediction by subgraph reasoning.
\newblock In \emph{International Conference on Machine Learning (ICML)}, 2020.

\bibitem[Toutanova \& Chen(2015)Toutanova and Chen]{toutanova2015observed}
Toutanova, K. and Chen, D.
\newblock Observed versus latent features for knowledge base and text
  inference.
\newblock In \emph{Proceedings of the 3rd Workshop on Continuous Vector Space
  Models and their Compositionality}, 2015.

\bibitem[Trouillon et~al.(2016)Trouillon, Welbl, Riedel, Gaussier, and
  Bouchard]{trouillon2016complex}
Trouillon, T., Welbl, J., Riedel, S., Gaussier, {\'E}., and Bouchard, G.
\newblock Complex embeddings for simple link prediction.
\newblock In \emph{International Conference on Machine Learning (ICML)}, 2016.

\bibitem[Vashishth et~al.(2020)Vashishth, Sanyal, Nitin, and
  Talukdar]{vashishthcomposition}
Vashishth, S., Sanyal, S., Nitin, V., and Talukdar, P.
\newblock Composition-based multi-relational graph convolutional networks.
\newblock In \emph{International Conference on Learning Representations
  (ICLR)}, 2020.

\bibitem[Wang et~al.(2023)Wang, Fu, Du, Gao, Huang, Liu, Chandak, Liu,
  Van~Katwyk, Deac, et~al.]{wang2023scientific}
Wang, H., Fu, T., Du, Y., Gao, W., Huang, K., Liu, Z., Chandak, P., Liu, S.,
  Van~Katwyk, P., Deac, A., et~al.
\newblock Scientific discovery in the age of artificial intelligence.
\newblock \emph{Nature}, 620\penalty0 (7972):\penalty0 47--60, 2023.

\bibitem[Xiao et~al.(2021)Xiao, Meng, Zhang, Wang, and Lin]{xiao2021training}
Xiao, M., Meng, Q., Zhang, Z., Wang, Y., and Lin, Z.
\newblock Training feedback spiking neural networks by implicit differentiation
  on the equilibrium state.
\newblock In \emph{Advances in Neural Information Processing Systems
  (NeurIPS)}, 2021.

\bibitem[Xiao et~al.(2022)Xiao, Meng, Zhang, He, and Lin]{xiao2022online}
Xiao, M., Meng, Q., Zhang, Z., He, D., and Lin, Z.
\newblock Online training through time for spiking neural networks.
\newblock In \emph{Advances in Neural Information Processing Systems
  (NeurIPS)}, 2022.

\bibitem[Xiao et~al.(2024)Xiao, Meng, Zhang, He, and Lin]{xiao2024hebbian}
Xiao, M., Meng, Q., Zhang, Z., He, D., and Lin, Z.
\newblock Hebbian learning based orthogonal projection for continual learning
  of spiking neural networks.
\newblock In \emph{International Conference on Learning Representations
  (ICLR)}, 2024.

\bibitem[Yan et~al.(2024)Yan, Meng, Xiao, Wang, and Lin]{yan2024sampling}
Yan, S., Meng, Q., Xiao, M., Wang, Y., and Lin, Z.
\newblock Sampling complex topology structures for spiking neural networks.
\newblock \emph{Neural Networks}, 172:\penalty0 106121, 2024.

\bibitem[Yan et~al.(2021)Yan, Ma, Gao, Tang, and Chen]{yan2021link}
Yan, Z., Ma, T., Gao, L., Tang, Z., and Chen, C.
\newblock Link prediction with persistent homology: An interactive view.
\newblock In \emph{International Conference on Machine Learning (ICML)}, 2021.

\bibitem[Yang et~al.(2015)Yang, Yih, He, Gao, and Deng]{yang2015embedding}
Yang, B., Yih, S. W.-t., He, X., Gao, J., and Deng, L.
\newblock Embedding entities and relations for learning and inference in
  knowledge bases.
\newblock In \emph{International Conference on Learning Representations
  (ICLR)}, 2015.

\bibitem[Yang et~al.(2017)Yang, Yang, and Cohen]{yang2017differentiable}
Yang, F., Yang, Z., and Cohen, W.~W.
\newblock Differentiable learning of logical rules for knowledge base
  reasoning.
\newblock In \emph{Advances in Neural Information Processing Systems
  (NeurIPS)}, 2017.

\bibitem[Yin et~al.(2021)Yin, Corradi, and Boht{\'e}]{yin2021accuratenmi}
Yin, B., Corradi, F., and Boht{\'e}, S.~M.
\newblock Accurate and efficient time-domain classification with adaptive
  spiking recurrent neural networks.
\newblock \emph{Nature Machine Intelligence}, 3\penalty0 (10):\penalty0
  905--913, 2021.

\bibitem[Zhang \& Chen(2018)Zhang and Chen]{zhang2018link}
Zhang, M. and Chen, Y.
\newblock Link prediction based on graph neural networks.
\newblock In \emph{Advances in Neural Information Processing Systems
  (NeurIPS)}, 2018.

\bibitem[Zhang et~al.(2024)Zhang, Xiao, Ji, Jiang, Lin, Zhou, and
  Lin]{zhang2024efficient}
Zhang, Z., Xiao, M., Ji, T., Jiang, Y., Lin, T., Zhou, X., and Lin, Z.
\newblock Efficient and generalizable cross-patient epileptic seizure detection
  through a spiking neural network.
\newblock \emph{Frontiers in Neuroscience}, 17:\penalty0 1303564, 2024.

\bibitem[Zheng et~al.(2022)Zheng, Lin, Zhao, and Shi]{zhengdance}
Zheng, H., Lin, H., Zhao, R., and Shi, L.
\newblock Dance of snn and ann: Solving binding problem by combining spike
  timing and reconstructive attention.
\newblock In \emph{Advances in Neural Information Processing Systems
  (NeurIPS)}, 2022.

\bibitem[Zhou et~al.(2021)Zhou, Li, Chen, Chandrasekaran, and
  Sanyal]{zhou2021temporal}
Zhou, S., Li, X., Chen, Y., Chandrasekaran, S.~T., and Sanyal, A.
\newblock Temporal-coded deep spiking neural network with easy training and
  robust performance.
\newblock In \emph{AAAI Conference on Artificial Intelligence (AAAI)}, 2021.

\bibitem[Zhu et~al.(2021)Zhu, Zhang, Xhonneux, and Tang]{zhu2021neural}
Zhu, Z., Zhang, Z., Xhonneux, L.-P., and Tang, J.
\newblock Neural bellman-ford networks: A general graph neural network
  framework for link prediction.
\newblock In \emph{Advances in Neural Information Processing Systems
  (NeurIPS)}, 2021.

\bibitem[Zhu et~al.(2022)Zhu, Peng, Li, Chen, Yu, and Luo]{zhu2022spiking}
Zhu, Z., Peng, J., Li, J., Chen, L., Yu, Q., and Luo, S.
\newblock Spiking graph convolutional networks.
\newblock In \emph{International Joint Conference on Artificial Intelligence
  (IJCAI)}, 2022.

\end{thebibliography}
\bibliographystyle{icml2024}

\newpage
\appendix
\onecolumn

\section{Training Spiking Neural Networks}\label{supp:sec:snn_training}

As introduced in \cref{sec:snn}, the models for membrane potential and current are described by the following equations:
\begin{equation}
    \small
    \tau_m\frac{du}{dt} = -(u(t)-u_{rest}) + R\cdot I(t),\quad u(t) < V_{th},
\end{equation}
\begin{equation}
    \small
    \tau_c\frac{dI_i}{dt} = -I_i(t) + \sum_j w_{ij} s_j(t - d_{ij}) + b_i,
\end{equation}
and the equivalent \ac{srm} formulation is:
\begin{equation}
    \small
    u_i(t) = u_{rest}+\sum_jw_{ij}\int_0^t\kappa(\tau-d_{ij})s_j(t-\tau) \mathrm{d}\tau + \int_0^t\nu(\tau)s_i(t-\tau) \mathrm{d}\tau.
\end{equation}

Let \( \mathcal{L} \) denote the loss based on the spikes of neurons. With the \ac{srm} formulation, the gradients for \( w_{ij} \) and \( d_{ij} \) can be calculated as follows:
\begin{equation}
    \frac{\partial \mathcal{L}}{\partial w_{ij}} = \int_0^T \delta_i(t) \frac{\partial s_i(t)}{\partial u_i(t)} \left(\int_0^t\kappa(\tau-d_{ij})s_j(t-\tau) \mathrm{d}\tau \right) \mathrm{d}t,
\end{equation}
\begin{equation}
    \frac{\partial \mathcal{L}}{\partial d_{ij}} = \int_0^T \delta_i(t) \frac{\partial s_i(t)}{\partial u_i(t)} w_{ij} \left(-\int_0^t\dot{\kappa}(\tau-d_{ij})s_j(t-\tau) \mathrm{d}\tau \right) \mathrm{d}t,
\end{equation}
where \( \delta_i(t) \) is the gradient for \( s_i(t) \) and can be recursively calculated by backpropagation through time as:
\begin{equation}
    \delta_i(t) = \frac{\partial \mathcal{L}}{\partial s_i(t)} + \int_t^T \left(\sum_{j}\delta_j(\tau) \frac{\partial s_j(\tau)}{\partial u_j(\tau)} \frac{\partial u_j(\tau)}{\partial s_i(t)} + \delta_i(\tau)\frac{\partial s_i(\tau)}{\partial u_i(\tau)}\frac{\partial u_i(\tau)}{\partial s_i(t)} \right) \mathrm{d}\tau,
\end{equation}
and \( \dot{\kappa}(\cdot) \) represents the derivative of \( \kappa(\cdot) \).

In practice, we simulate \acp{snn} using the discrete computational form of the \ac{lif} model:
\begin{equation}
    \small
    \left\{
    \begin{aligned}
        &I_i[t + 1] = \exp\left(-\frac{\Delta \tau}{\tau_c}\right) I_i[t] + \alpha \left(\sum_j w_{ij} s_j[t - d_{ij}] + b_i\right),\\
        &u_i[t + 1] = \exp\left(-\frac{\Delta \tau}{\tau_m}\right) u_i[t] (1 - s_i[t]) + I_i[t + 1],\\
        &s_i[t + 1] = H(u_i[t+1] - V_{th}).\\
    \end{aligned}
    \right.
\end{equation}
The gradients of \( \delta_i(t) \) and \( \frac{\partial \mathcal{L}}{\partial w_{ij}} \) can be calculated using the standard backpropagated automatic differentiation framework in deep learning libraries, based on the above formulation. The spiking function is non-differentiable, and \( \frac{\partial s_i[t]}{\partial u_i[t]} \) can be replaced by a surrogate derivative~\citep{shrestha2018slayer}. We consider the derivative of the sigmoid function:
\begin{equation}
    \small
    \frac{\partial s}{\partial u}=\frac{1}{a_1}\frac{e^{(V_{th}-u)/a_1}}{(1+e^{(V_{th}-u)/a_1})^2},
\end{equation}
where we take \( a_1=0.25 \).

The automatic differentiation of the above formulation cannot directly handle $\frac{\partial \mathcal{L}}{\partial d_{ij}}$. We rewrite it in the discrete setting as:
\begin{equation}
    \small
    \frac{\partial \mathcal{L}}{\partial d_{ij}}= \sum_t\delta_i[t] \frac{\partial s_i[t]}{\partial u_i[t]} w_{ij} \left(-\sum_{\tau=0}^{t-1} \dot{\kappa}[\tau-d_{ij}]s_j[t-1-\tau]\right).
\end{equation}
We can integrate this into the automatic differentiation by tracking the trace ${tr}_{ij}[t]=-\sum_{\tau=0}^t \dot{\kappa}[\tau-d_{ij}]s_j[t-\tau]$ and calculating gradients based on it and the error backpropagated to $s_j[t - d_{ij}]$. In the discrete setting, $d_{ij}$ should be an integer index. We quantize it in the forward simulation and calculate gradients using the straight-through-estimator.

As described in \cref{sec:snn}, we consider $\tau_c=\tau_m, R=e$ and the input kernel is $\kappa(\tau)=\frac{e\tau}{\tau_m}\exp\left(-\frac{\tau}{\tau_m}\right)$ for $\tau\geq0$ and $\kappa(\tau)=0$ for $\tau<0$. Then, $\dot{\kappa}(\tau)=\frac{e}{\tau_m} \left(1 - \frac{\tau}{\tau_m}\right) \exp\left(-\frac{\tau}{\tau_m}\right)$ for $\tau\geq0$. In the discrete setting of the \ac{lif} model, $\kappa$ is better described as $\kappa[\tau]=\alpha (\tau+1)\exp(-\frac{\tau}{\tau_m / \Delta \tau}), \tau\geq 0$ (where $\alpha=\frac{e}{\tau_m / \Delta \tau}$). Correspondingly, we take $\dot{\kappa}[\tau]=\alpha \left(1 - \frac{\tau+1}{\tau_m / \Delta \tau}\right) \exp\left(-\frac{\tau}{\tau_m / \Delta \tau}\right)$ and calculate the trace ${tr}_{ij}$ based on it.

\section{Proof of Proposition 3.1}\label{supp:sec:proof}

\begin{proposition}
Katz Index, Personalized PageRank, and Graph Distance can be solved by \ac{grsnn} under specific settings.
\end{proposition}

\begin{proof}

We first introduce more details of Katz Index, Personalized PageRank, and Graph Distance. As described in \cref{sec:snn_path_formulation}, traditional path-based algorithms for graphs calculate the pair representation between nodes $x, y$ by considering paths from $x$ to $y$, and this can be formulated as a generalized accumulation of path representations (denoted as $\otimes$) with a commutative summation operator (denoted as $\oplus$):
\begin{equation}
\small
    \mathbf{h}^q(x, y) = \bigoplus_{P\in\mathcal{P}_{xy}} \left(\bigotimes_{i=1}^{\vert P\vert}\mathbf{v}^q(e_i)\right),
\end{equation}
where $\mathcal{P}_{xy}$ is the set of paths from $x$ to $y$, $e_i$ is the $i$-th edge on a path $P$, and $\mathbf{v}^q(e_i)$ is the representation of the edge (\eg, the transition probability of this edge). Katz Index is a path formulation with $\oplus=+, \otimes=\times, \mathbf{v}^q(e)=\beta$, Personalized PageRank is with $\oplus=+, \otimes=\times, \mathbf{v}^q(e)=1/d_{out}(z)$ (where $d_{out}(z)$ is the output degree of the start node $z$ of edge $e$), and Graph Distance is with $\oplus=\min, \otimes=+, \mathbf{v}^q(e)=1$.

We examine these three distinct settings:

\textbf{(1) Graph Distance:} 
In this setting, each graph node is assigned one spiking neuron, and neurons are connected if there exists a graph edge between them, with all synaptic weights and thresholds set to $1$. Consequently, each input spike to a neuron will trigger an output spike. The synaptic delay of each edge is set as the corresponding positive graph edge length, allowing the propagation of spikes along edges to accumulate edge length into time. By initiating a spike from the source node at time $0$, \ac{grsnn} propagates spikes throughout the network, and the time to the first spike of each node represents the shortest distance to the source node. Utilizing the decoding function of the spike train from the target node as the first spiking time allows us to compute the graph distance.

\textbf{(2) Katz Index:} 
The Katz Index necessitates the accumulative multiplication of edge representations. By applying the log operation, this multiplication can be transformed into accumulation. For an edge representation $\beta \in (0, 1)$ of Katz Index, corresponding to an attenuation factor, the synaptic delay is set as $d = -\log\beta$ (potentially scaled). For a spiking time $t$, $10^{-t}$ represents the accumulative multiplication of edge representations in the path. To sum over all paths, the number of paths during spike propagation must be maintained. A single spiking neuron is insufficient for this task as it will only generate one output spike when multiple paths simultaneously propagate to the same node. This limitation can be addressed by employing multiple spiking neurons, assigning $N$ spiking neurons to each graph node, with thresholds set as $1,2,\cdots,N$. Neurons connected by graph edges have synaptic weights of $1$ and delays as described above. The time and number of spikes of each node correspond to different paths from the source node. After sufficient propagation time, the decoding function of the spike train from the target node is defined as $D(\mathbf{s}(t))=\sum_{\tau}10^{-\tau}\left(\sum_i s_i[\tau]\right)$, enabling the computation of the Katz Index.

\textbf{(3) Personalized PageRank:} 
This is analogous to the Katz Index, with the edge representation being the transition probability $1/d_{out}(z) \in (0, 1)$. The synaptic delay is similarly set as $d = -\log(1/d_{out}(z))$ (or with a scale). Thus, Personalized PageRank can be computed similarly to the Katz Index.

\end{proof}

\begin{remark}
The crux of the proof revolves around the construction of appropriate synaptic delays and decoding functions. As illustrated in the construction, distinct temporal coding methods naturally arise for varying path formulations. In many scenarios, the significance of edge representations in knowledge graphs can be interpreted as learnable probabilities, making the accumulative multiplication setting (as in Katz Index and Personalized PageRank) particularly advantageous. This results in the adoption of temporal coding in our experiments in the main text, assigning different weights to different spikes, represented as $D(\mathbf{s}_y^q(t))=\frac{\sum_{\tau}\lambda^{\tau}\mathbf{s}_y^q[\tau]}{\sum_{\tau}\lambda^\tau}$, except a constant factor. A notable distinction is that, instead of a straightforward summation across different neurons, we derive the pair representation as a vector of different neurons. Subsequently, the likelihood is computed using a learnable function $g$, aligning with the prevalent approaches in graph reasoning methods (refer to \cref{sec:implementation_details}). This approach also serves as a broader generalization of the formulation in the construction.
\end{remark}

\section{Related Work}

\textbf{Spiking Neural Networks}\quad Recent works mainly study SNNs as energy-efficient alternatives to ANNs by converting ANNs to SNNs for object recognition~\citep{rueckauer2017conversion,stockl2021optimized,li2021free,meng2022trainingnn} and natural language classification~\citep{lv2023spiking}, or direct training SNNs (with surrogate gradients or other methods) for audio or visual perception~\citep{shrestha2018slayer,Fang_2021_ICCV,xiao2021training,meng2022training,xiao2024hebbian}, time series classification~\citep{yin2021accuratenmi,rao2022long}, seizure detection~\citep{zhang2024efficient}, graph classification~\citep{zhu2022spiking,li2023scaling}, etc. Most of them focus on spike counts and hardly leverage the important temporal dimension. Some works explore temporal encoding for single neurons~\citep{mostafa2017supervised,comsa2020temporal,zhou2021temporal,stockl2021optimized}, or utilizing spiking time for feature binding~\citep{zhengdance}, but how synaptic delay with temporal coding at the network level can be systematically utilized is rarely considered. \citet{patino2023empirical} and \citet{hammouamri2024learning} learn delays for time series tasks, and \citet{yan2024sampling} learn delays in network structure design, but they not consider temporal coding for graph tasks. 
Some works attempt to use SNNs for relational reasoning in knowledge graphs with entity embedding based on spiking times~\citep{dold2021spike,dold2022relational} or population coding combined with reward-modulated STDP~\citep{fang2022brain}. They do not consider reasoning paths with synaptic delay and temporal coding, and are limited in inductive generalization and interoperability considering the entity embedding method as well as poor performance in large knowledge graphs. 
Differently, our novel method is the first to demonstrate the advantage of delays to represent relations with promising performance on real transductive and inductive (knowledge) graphs.

\textbf{Graph Reasoning }\quad Graph link prediction is a fundamental graph reasoning task, typically in the context of knowledge graph reasoning. Popular methods include three paradigms: path-based, embedding, and graph neural networks~\citep{zhu2021neural}. Path-based methods predict links based on paths from the source node to the target node, \eg, the weighted count of paths in homogeneous graphs~\citep{katz1953new,page1998pagerank,liben2007link} or paths with learned probabilities or representations in knowledge graphs~\citep{lao2010relational,yang2017differentiable,sadeghian2019drum}. 
Embedding methods learn representations for each node and edge which preserve the structure of the graph~\citep{perozzi2014deepwalk,tang2015line,grover2016node2vec,bordes2013translating,yang2015embedding,sunrotate}. 
They rely on entities and cannot perform inductive reasoning. GNNs perform message passing between nodes for reasoning based on the learned node or edge representations. For knowledge graphs, R-GCN~\citep{schlichtkrull2018modeling} and CompGCN~\citep{vashishthcomposition} propagate over all entities with different message functions, while GraIL~\citep{teru2020inductive} propagates in an extracted subgraph. NBFNet~\citep{zhu2021neural} proposes a framework to integrate path formulation and graph neural networks, achieving state-of-the-art results with \acp{gnn}. Different from these works, we focus on exploring SNNs with spiking time.

\section{Datasets statistics}\label{supp:sec:dataset_statistics}

FB15k-237~\citep{toutanova2015observed} is a refined knowledge graph link prediction dataset derived from FB15k. It is meticulously curated to ensure that the test and evaluation datasets are devoid of inverse relation test leakage. Similarly, WN18RR~\citep{dettmers2018convolutional} is another knowledge graph link prediction dataset, formulated from WN18 (a subset of WordNet), maintaining the integrity by avoiding inverse relation test leakage. 

For the conventional transductive knowledge graph completion setting, the datasets exhibit varying quantities of entities, relations, and relation triplets across the train, validation, and test sets, as detailed in \cref{supp:table:stat_transductive_kg}. In the context of the standard inductive relation prediction setting, the statistical breakdown for different splits is depicted in \cref{supp:table:stat_inductive_kg}. 

Additionally, Cora, Citeseer, and PubMed~\citep{sen2008collective} serve as homogeneous citation graphs, with their respective statistics outlined in \cref{supp:table:stat_homogeneous}.

\begin{table}[ht]
    \centering
    \small
    \caption{\textbf{Transductive Knowledge Graph Completion Statistics for FB15k-237 and WN18RR.}}
    \label{supp:table:stat_transductive_kg}
    \begin{tabular}{cccccc}
         \toprule
         \multirow{2}{*}{Dataset} & \multirow{2}{*}{\#Entity} & \multirow{2}{*}{\#Relation} & \multicolumn{3}{c}{\#Triplet} \\
         & & & \#Train & \#Validation & \# Test\\
         \midrule
         FB15k-237~\citep{toutanova2015observed} & 14,541 & 237 & 272,115 & 17,535 & 20,466\\
         WN18RR~\citep{dettmers2018convolutional} & 40,943 & 11 & 86,835 & 3,034 & 3,134\\
         \bottomrule
    \end{tabular}
\end{table}

\begin{table}[t!]
    \centering
    \small
    \caption{\textbf{Inductive Relation Prediction Statistics for FB15k-237 and WN18RR.}}
    \label{supp:table:stat_inductive_kg}
    \resizebox{\linewidth}{!}{%
        \begin{tabular}{cccccccccccc}
             \toprule
             \multirow{2}{*}{Dataset \& Split} & & \multirow{2}{*}{\#Relation} & \multicolumn{3}{c}{Train} & \multicolumn{3}{c}{Validation} & \multicolumn{3}{c}{Test} \\
             & & & \#Entity & \#Query & \#Fact & \#Entity & \#Query & \#Fact & \#Entity & \#Query & \#Fact\\
             \midrule
             \multirow{4}{*}{FB15k-237~\citep{teru2020inductive}} & v1 & 180 & 1,594 & 4,245 & 4,245 & 1,594 & 489 & 4,245 & 1,093 & 205 & 1,993\\
             & v2 & 200 & 2,608 & 9,739 & 9,739 & 2,608 & 1,166 & 9,739 & 1,660 & 478 & 4,145\\
             & v3 & 215 & 3,668 & 17,986 & 17,986 & 3,668 & 2,194 & 17,986 & 2,501 & 865 & 7,406\\
             & v4 & 219 & 4,707 & 27,203 & 27,203 & 4,707 & 3,352 & 27,203 & 3,051 & 1,424 & 11,714\\
             \midrule
             \multirow{4}{*}{WN18RR~\citep{teru2020inductive}} & v1 & 9 & 2,746 & 5,410 & 5,410 & 2,746 & 630 & 5,410 & 922 & 188 & 1,618\\
             & v2 & 10 & 6,954 & 15,262 & 15,262 & 6,954 & 1,838 & 15,262 & 2,757 & 441 & 4,011\\
             & v3 & 11 & 12,078 & 25,901 & 25,901 & 12,078 & 3,097 & 25,901 & 5,084 & 605 & 6,327\\
             & v4 & 9 & 3,861 & 7,940 & 7,940 & 3,861 & 934 & 7,940 & 7,084 & 1,429 & 12,334\\
             \bottomrule
        \end{tabular}%
    }%
\end{table}

\begin{table}[ht]
    \centering
    \small
    \caption{\textbf{Homogeneous Graph Link Prediction Statistics for Cora, CiteSeer, and PubMed.}}
    \label{supp:table:stat_homogeneous}
    \begin{tabular}{ccccc}
         \toprule
         \multirow{2}{*}{Dataset} & \multirow{2}{*}{\#Node} & \multicolumn{3}{c}{\#Edge} \\
         & & \#Train & \#Validation & \# Test\\
         \midrule
         Cora~\citep{sen2008collective} & 2,708 & 4,614 & 271 & 544\\
         CiteSeer~\citep{sen2008collective} & 3,327 & 4,022 & 236 & 474\\
         PubMed~\citep{sen2008collective} & 19,717 & 37,687 & 2,216 & 4,435\\
         \bottomrule
    \end{tabular}
\end{table}

\section{More Implementation and Experimental Details}

\subsection{Link Prediction Detail}\label{supp:sec:implementation_details}
In line with prevalent practices for link prediction, the objective is to ascertain the likelihood of a triplet $(x, q, y)$, consisting of the source node, query relation, and target node. 
Consistent with prior studies~\citep{zhu2021neural}, we employ a feed-forward neural network $g$ to estimate the conditional likelihood of the tail entity $y$, predicated on the head entity $x$ and query $q$, utilizing the pair representation $\mathbf{h}^q(x, y)$, formulated as $p(y \vert x, q)=\sigma(g(\mathbf{h}^q(x, y); \mathbf{r}^q))$, where $\sigma$ denotes the sigmoid function. Analogously, the conditional likelihood of the head entity $x$, contingent upon $y$ and $q$, is deduced as $p(x \vert y, q^{-1})=\sigma(g(\mathbf{h}^{q^{-1}}(y, x); \mathbf{r}^{q^{-1}}))$, with $q^{-1}$ representing the inverted relation. 
In the scenario of undirected graphs, the representations undergo symmetrization, resulting in $p(x, q, y)=\sigma(g(\mathbf{h}^q(x, y) + \mathbf{h}^q(y, x); \mathbf{r}^q))$. 
Adhering to established methodologies, a two-layer \ac{mlp} with ReLU activation is utilized for $g$. It is noteworthy that this configuration is also conducive to implementation via a spiking \ac{mlp}, given the facile conversion of the ReLU function to spiking neurons, achievable through rate or temporal coding~\citep{rueckauer2017conversion,stockl2021optimized}. \cref{supp:sec:spiking_head} also studies directly training a spiking \ac{mlp} for $g$ and the results remain about the same.

In short, the procedure of our model to deduce a triplet $(x, q, y)$ commences with the propagation of spike trains across the graph to secure the pair representation $\mathbf{h}^q(x, y)$, and subsequently, the likelihood score is computed by $g$, predicated on $\mathbf{h}^q(x, y)$. When provided with the head entity $x$ and the query relation $r$, the model is capable of concurrently computing pair representations and scores for all conceivable tail entities during the forward propagation of \acp{snn}. The overarching procedure aligns with the conventional graph reasoning paradigm, with our primary focus being on the pivotal step of acquiring the pair representation through \ac{snn} propagation.

Regarding the training procedure, we adhere to the methodologies of preceding works~\citep{bordes2013translating,sunrotate,zhu2021neural}, generating negative samples by corrupting one entity in a positive triplet. The training objective is formulated to minimize the negative log-likelihood of both positive and negative triplets:
\begin{equation}
    \small
    \mathcal{L} = -\log p(x, q, y) - \sum_{i=1}^m \frac{1}{m} \log (1 - p(x_i', q, y_i')),
\end{equation}
where $m$ is the number of negative samples for each positive one, and $(x_i', q, y_i')$ denotes the $i$-th negative sample.

\subsection{Task Details}\label{supp:sec:task_details}

\begin{figure*}[ht]
    \centering
    \includegraphics[width=\linewidth]{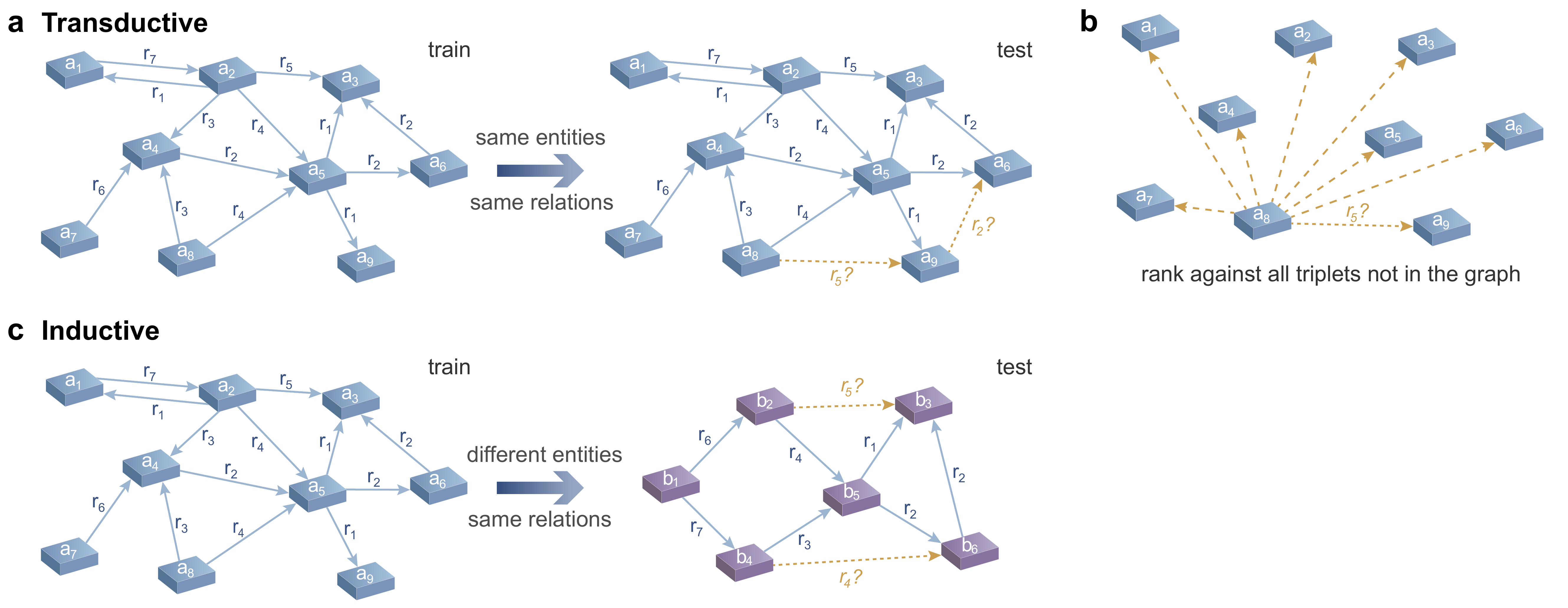}
    \caption{\textbf{Illustration of task details.}  (a) Depiction of the transductive knowledge graph completion process. (b) Illustration of the filtered ranking protocol used to rank the test triplet $(x, q, y)$ against all negative triplets absent from the graph. The triplets $(x', q, y)$ are not shown here for clarity. (c) Illustration of the inductive setting of relation prediction.}
    \label{supfig:task_details}
\end{figure*}

We illustrate the tasks of transductive knowledge graph completion and inductive relation prediction in \cref{supfig:task_details}. For homogeneous graph link prediction, it is similar to transductive knowledge graph completion except that there is only one relation type in homogeneous graphs, \ie, the existence of the edge.

\subsection{Experimental Details} \label{supp:sec:experimental_details}

\paragraph{Datasets and preprocessing}
We assess our model across various tasks including transductive knowledge graph completion, inductive knowledge graph relation prediction, and homogeneous graph link prediction. For knowledge graphs, we employ the widely recognized FB15k-237~\citep{toutanova2015observed} and WN18RR~\citep{dettmers2018convolutional}, adhering to the standard transductive~\citep{toutanova2015observed,dettmers2018convolutional} and inductive splits~\citep{teru2020inductive}. For homogeneous graphs, we utilize Cora, Citeseer, and PubMed~\citep{sen2008collective}. 

In evaluating knowledge graph completion, we adhere to the prevalent filtered ranking protocol~\citep{bordes2013translating}, ranking a test triplet $(x, q, y)$ against all negative triplets $(x, q, y')$ or $(x', q, y)$ absent in the graph (considering the likelihood score). We report \ac{mr}, \ac{mrr}, and HITS at N. For inductive knowledge graph relation prediction, we align with the previous practice~\citep{teru2020inductive}, drawing 50 negative triplets for each positive one using the aforementioned filtered ranking and report HITS@10. In the context of homogeneous graph link prediction, we follow the approaches of~\citet{kipf2016variational}, contrasting the positive edges with an equivalent number of negative edges, and report \ac{auroc} and \ac{ap}. The distribution of edges in train/valid/test is maintained at a ratio of 85:5:10, aligning with common practice. The specifics and statistics related to the datasets are available in \cref{supp:sec:dataset_statistics}.

Regarding data preprocessing, we adhere to the methodologies of prior works~\citep{yang2017differentiable,sadeghian2019drum,kipf2016variational}. In knowledge graphs, each triplet $(x, q, y)$ is augmented with a reversed triplet $(y, q^{-1}, x)$. In homogeneous graphs, each node is augmented with a self-loop. Additionally, we follow~\citet{zhu2021neural} to exclude edges directly connecting query node pairs during the training phase for the transductive setting of FB15k-237 and homogeneous graphs.

\paragraph{Models and training}
Given the substantial computational expense associated with simulating \acp{snn} over a long time, our primary simulations involve $T=10$ discrete time steps for \acp{snn}. The hyperparameters for \acp{snn} are designated as $\tau_m / \Delta \tau=4, V_{th}=2$, with the discrete delay bound $\beta=4$, and $\lambda=0.95$ for the decoding function. This can correspond to $\tau_m=\tau_c=20 ms$ with discretization interval $\Delta \tau=5 ms$ and a total simulation time $T\times\Delta\tau=50ms$ for \acp{snn}. For experiments analyzing temporal discretization steps in \cref{supp:sec:impact_discretization}, hyperparameters are adjusted relative to the discrete step; for instance, for $T=5$, we assign $\tau_m / \Delta \tau=2, \beta=2, \lambda=0.9$ (corresponding to $\Delta \tau=10ms$), and for $T=20$, we designate $\tau_m / \Delta \tau=8, \beta=8, \lambda=0.97$ (corresponding to $\Delta \tau=2.5ms$). Each graph node is represented by $n=32$ spiking neurons by default. No normalization or other modifications are applied, and for models on FB15k-237, a linear scale of $0.1$ is applied post the linear transformation $\mathbf{W}$. 

As for the two baseline \ac{snn} models that we compare in \cref{sec:experiments} to elucidate the superiority of synaptic delay, the first model abstains from encoding edge relations, and the delay $\mathbf{d}_r^q$ in \cref{eq:grsnn} is not taken into account, \ie, it is assigned a value of zero. The second model opts for encoding relations through synaptic weight instead of synaptic delay. We modify $\mathbf{s}^q_k[t - \mathbf{d}^q_{r}]$ in \cref{eq:grsnn} to $\mathbf{w}_r^q\odot\mathbf{s}^q_k[t]$ (where $\mathbf{w}_r^q$ is defined analogously to $\mathbf{d}^q_{r}$ but devoid of the sigmoid function and bound scale, and $\mathbf{w}_r^q$ can be amalgamated into $\mathbf{W}$ to formulate the entire synaptic weight). This alteration aligns with the DistMult message function utilized in prior works to multiply messages with edge representations~\citep{zhu2021neural}.

For \textit{GRSNN+}, we apply layer normalization (LN) after the linear transformation of the aggregated messages as in many \acp{gnn}, and encode relations in both synaptic delay and synaptic weight, \ie, the messages are $\mathbf{w}_r^q\odot\mathbf{s}^q_k[t - \mathbf{d}^q_{r}]$. For FB15k-237, we further adopt the principal neighborhood aggregation (PNA) as the aggregation function instead of summation, which is a major component for the high performance of NBFNet~\citep{zhu2021neural}. We show that by integrating these \ac{gnn} tricks, \ac{grsnn} can also achieve a better performance.

All models are trained utilizing the Adam optimizer over 20 epochs. The learning rate is $2e-3$ for transductive settings (knowledge graph completion and homogeneous graph link prediction) and $5e-3$ for inductive settings. The batch size is 32 (30 for transductive FB15k-237), achieved by accumulating gradients across several iterations with smaller mini-batches each iteration.

The ratio of negative samples is configured to 256 for FB15k-237 and WN18RR in the transductive setting and 50 in the inductive setting to align more closely with testing conditions, while it is established as $1$ for homogeneous graphs, adhering to previous studies. The temperature in self-adversarial negative sampling is determined to be 0.5 and 1 for FB15k-237 and WN18RR, respectively. Model selection is based on validation performance, with \ac{mrr} serving as the criterion for knowledge graphs and \ac{auroc} for homogeneous graphs.

Our code implementation leverages the PyTorch framework, and experimental evaluations are executed on one or two NVIDIA GeForce RTX 3090 GPUs.

\paragraph{Details of theoretical energy estimation}
For theoretical inference operation counts and energy estimations, we consider the scenario where neural network models are deployed and mapped directly to individual neurons and synapses. This scenario aligns with the principles of neuromorphic computing and hardware~\citep{davies2018loihi,pei2019towards,rao2022long}, facilitating in-memory computation and minimizing energy-consuming memory exchanges. Our theoretical analysis predominantly centers on the operations of neurons and synapses, omitting additional hardware-related costs such as memory access.

For the spiking model, the estimated synaptic operations are given by $T\times n^2\times fr \times \vert\mathcal{E}\vert$, where $T$ represents the discrete time step, $n$ is the number of neurons allocated per graph node, $fr$ denotes the spike rate, and $\vert\mathcal{E}\vert$ is the count of graph edges. This calculation corresponds to the quantity of synaptic operations instigated by spikes, culminating in an accumulation (addition) operation of post-synaptic current (or membrane potential). Additionally, accounting for neuron dynamics, there will be $T\times n \times \vert\mathcal{V}\vert$ addition operations for the bias term, $T\times n \times \vert\mathcal{V}\vert$ addition operations for the accumulation of membrane potential with current, and $2T\times n \times \vert\mathcal{V}\vert$ multiplication operations due to the leakage of current and membrane potential, where $\vert\mathcal{V}\vert$ represents the number of graph nodes. The computational cost associated with spike generation and reset is omitted in this estimation. Consequently, the total operations involve $2T\times n \times \vert\mathcal{V}\vert$ multiplications and $T\times n^2\times fr \times \vert\mathcal{E}\vert + 2T\times n \times \vert\mathcal{V}\vert$ additions.

For the non-spiking counterpart, assuming the replacement of spiking neurons with conventional artificial neurons and disregarding the computational cost of the activation function, the synaptic operations would involve $T\times n^2\times \vert\mathcal{E}\vert$ \ac{mac} operations (multiplication + addition), along with $T\times n \times \vert\mathcal{V}\vert$ addition operations for the bias term. Thus, the total operations would encompass $T\times n^2\times \vert\mathcal{E}\vert$ multiplications and $T\times n^2\times (\vert\mathcal{E}\vert + \vert\mathcal{V}\vert)$ additions.

\paragraph{Costs of synaptic delay}
We consider the Ring Buffer for potential synaptic delay implementation as analyzed in \citet{patino2023empirical}, which is commonly used by digital neuromorphic platforms. The memory overhead of the ring buffer is the number of neurons multiplied by the maximum synaptic delay $M_d$, and the energy overhead is equal to one extra neural accumulation per time step for each neuron \citep{patino2023empirical}. Then, the additional memory overhead (words) is $n\times \vert\mathcal{V}\vert\times M_d$ and the additional energy overhead is $T\times n\times \vert\mathcal{V}\vert\times E_{AC}$. Consider the original memory overhead $n\times \vert\mathcal{V}\vert\times 2 + n^2\times \vert\mathcal{E}\vert$ (neuron states + synapses) and the original energy overhead $2T\times n \times \vert\mathcal{V}\vert\times E_{MAC} + (T\times n^2\times fr \times \vert\mathcal{E}\vert + 2T\times n \times \vert\mathcal{V}\vert)\times E_{AC}$ (analyzed above), the additional overhead is small because the number of synapses is much larger than the number of neurons ($n^2\times \vert\mathcal{E}\vert \gg n\times \vert\mathcal{V}\vert $) in our settings. Specifically, on the test set of FB15k-237, the originally analyzed energy is estimated as 1.337 mJ, while the additional energy for synaptic delay is estimated as 0.004 mJ, which is marginal.

\paragraph{Visualization of reasoning paths}
The methodology for visualizing reasoning paths in \cref{supp:sec:interpretability} is elucidated below. The interpretation of reasoning is predicated on the significance of paths to the concluding prediction score. According to \citet{zhu2021neural}, this significance or importance can be computed by the gradient of the prediction with respect to the paths, based on the local 1st-order Taylor expansion, and the path importance can be approximated by summing the importance of the edges in the path. This edge importance is computed using automatic differentiation. Specifically, during the forward procedure, the variable of edge weight (initialized to 1) is multiplied to the message transmitted through this edge (\ie, the delayed spikes, with 1 representing a spike and 0 representing no spike). Only when a spike is present will there be a gradient for this variable during backpropagation. Subsequently, during backpropagation, this variable accumulates the gradients of all neurons at every time step, representing the edge importance. 

For the non-differentiable spiking operation, a distinct surrogate gradient is employed for backpropagation. If the membrane potential $u$ is below the threshold, the gradient is set to 0, as there is no output spike influencing other neurons. Conversely, if the membrane potential surpasses the threshold, the gradient is set as $1/u$, normalizing the contribution of inputs to the output based on the membrane potential, as the gradient of the output is for spike $1$. 

The top-k path importance is thus analogous to the top-k longest paths when considering edge importance. We adopt a beam search, as suggested by \citet{zhu2021neural}, to identify these paths. It is crucial to note that this method provides only a rough approximation, and future research may explore more refined interpretative approaches.

\section{More Results and Detailed Values}\label{supp:sec:more_results}

\subsection{Detailed Values of Main Results}\label{supp:sec:detailed_values}
In this section, we furnish detailed results for various experiments. The comprehensive result values for transductive knowledge graph completion are presented in \cref{supp:table:transductive_knowledge_graph}. For inductive relation prediction, the detailed results can be referred to in \cref{supp:table:inductive_knowledge_graph}. Lastly, the exhaustive result values for homogeneous graph link prediction are available in \cref{supp:table:homogeneous_graph}.

\begin{table}[t!]
    \centering
    \small
    \caption{\textbf{Detailed Results for Transductive Knowledge Graph Completion.} Lower values are preferable for MR, while higher values are preferable for MRR, HITS@1, HITS@3, and HITS@10. *SpikTE is an embedding method based on spiking neurons.}
    \label{supp:table:transductive_knowledge_graph}
    \resizebox{\linewidth}{!}{%
        \begin{tabular}{cccccccccccc}
            \multicolumn{12}{c}{(b)}\\
            \toprule
            \multirow{2}{*}{Class} & \multirow{2}{*}{Method} & \multicolumn{5}{c}{FB15k-237} & \multicolumn{5}{c}{WN18RR} \\
            & & MR$\downarrow$ & MRR$\uparrow$ & H@1$\uparrow$ & H@3$\uparrow$ & H@10$\uparrow$ & MR$\downarrow$ & MRR$\uparrow$ & H@1$\uparrow$ & H@3$\uparrow$ & H@10$\uparrow$\\
            \midrule
            \multirow{3}{*}{Path-based} & Path Ranking~\citep{lao2010relational} & 3521 & 0.174 & 0.119 & 0.186 & 0.285 & 22438 & 0.324 & 0.276 & 0.360 & 0.406\\
            & NeuralLP~\citep{yang2017differentiable} & - & 0.240 & - & - & 0.362 & - & 0.435 & 0.371 & 0.434 & 0.566\\
            & DRUM~\citep{sadeghian2019drum} & - & 0.343 & 0.255 & 0.378 & 0.516 & - & 0.486 & 0.425 & 0.513 & 0.586\\
            \midrule
            \multirow{6}{*}{Embeddings} & TransE~\citep{bordes2013translating} & 357 & 0.294 & - & - & 0.465 & 3384 & 0.226 & - & - & 0.501\\
            & DistMult~\citep{yang2015embedding} & 254 & 0.241 & 0.155 & 0.263 & 0.419 & 5110 & 0.43 & 0.39 & 0.44 & 0.49\\
            & ComplEx~\citep{trouillon2016complex} & 339 & 0.247 & 0.158 & 0.275 & 0.428 & 5261 & 0.44 & 0.41 & 0.46 & 0.51\\
            & RotatE~\citep{sunrotate} & 177 & 0.338 & 0.241 & 0.375 & 0.533 & 3340 & 0.476 & 0.428 & 0.492 & 0.571\\
            & LowFER~\citep{amin2020lowfer} & - & 0.359 & 0.266 & 0.396 & 0.544 & - & 0.465 & 0.434 & 0.479 & 0.526\\
            & SpikTE*~\citep{dold2022relational} & - & 0.21 & 0.13 & 0.23 & - & - & - & - & - & -\\
            \midrule
            \multirow{4}{*}{\acp{gnn}} & RGCN~\citep{schlichtkrull2018modeling} & 221 & 0.273 & 0.182 & 0.303 & 0.456 & 2719 & 0.402 & 0.345 & 0.437 & 0.494\\
            & GraIL~\citep{teru2020inductive} & 2053 & - & - & - & - & 2539 & - & - & - & -\\
            & CompGCN~\citep{vashishthcomposition} & 197 & 0.355 & 0.264 & 0.390 & 0.535 & 3533 & 0.479 & 0.443 & 0.494 & 0.546\\
            & NBFNet~\citep{zhu2021neural} & 114 & 0.415 & 0.321 & 0.454 & 0.599 & 636 & 0.551 & 0.497 & 0.573 & 0.666\\
            \midrule
            \multirow{2}{*}{\acp{snn}} & \textbf{\ac{grsnn} (ours)} & 139 & 0.368 & 0.275 & 0.407 & 0.551 & 720 & 0.508 & 0.455 & 0.528 & 0.616\\
            & \textbf{\textit{GRSNN+} (ours)} & 132 & 0.393 & 0.301 & 0.431 & 0.572 & 610 & 0.532 & 0.478 & 0.557 & 0.637\\
            \bottomrule
        \end{tabular}%
    }%
\end{table}

\begin{table*}[t!]
    \centering
    \small
    \caption{\textbf{Detailed Results for Inductive Relation Prediction (HITS@10).} v1-v4 correspond to the four standard versions of inductive splits.}
    \label{supp:table:inductive_knowledge_graph}
        \begin{tabular}{cccccccccc}
            \toprule
            \multirow{2}{*}{Class} & \multirow{2}{*}{Method} & \multicolumn{4}{c}{FB15k-237} & \multicolumn{4}{c}{WN18RR} \\
            & & v1 & v2 & v3 & v4 & v1 & v2 & v3 & v4\\
            \midrule
            \multirow{3}{*}{Path-based} & NeuralLP~\citep{yang2017differentiable} & 0.529 & 0.589 & 0.529 & 0.559 & 0.744 & 0.689 & 0.462 & 0.671\\
            & DRUM~\citep{sadeghian2019drum} & 0.529 & 0.587 & 0.529 & 0.559 & 0.744 & 0.689 & 0.462 & 0.671\\
            & RuleN~\citep{meilicke2018fine} & 0.498 & 0.778 & 0.877 & 0.856 & 0.809 & 0.782 & 0.534 & 0.716\\
            \midrule
            \multirow{2}{*}{\acp{gnn}} & GraIL~\citep{teru2020inductive} & 0.642 & 0.818 & 0.828 & 0.893 & 0.825 & 0.787 & 0.584 & 0.734\\
            & NBFNet~\citep{zhu2021neural} & 0.834 & 0.949 & 0.951 & 0.960 & 0.948 & 0.905 & 0.893 & 0.890\\
            \midrule
            \acp{snn} & \textbf{\ac{grsnn} (ours)} & 0.852 & 0.957 & 0.958 & 0.958 & 0.943 & 0.892 & 0.906 & 0.888\\
            \bottomrule
        \end{tabular}%
\end{table*}

\begin{table*}[t!]
    \centering
    \small
    \caption{\textbf{Detailed Results for Homogeneous Graph Link Prediction.}}
    \label{supp:table:homogeneous_graph}
        \begin{tabular}{cccccccc}
            \toprule
            \multirow{2}{*}{Class} & \multirow{2}{*}{Method} & \multicolumn{2}{c}{Cora} & \multicolumn{2}{c}{Citeseer} & \multicolumn{2}{c}{PubMed} \\
            & & AUROC$\uparrow$ & AP$\uparrow$ & AUROC$\uparrow$ & AP$\uparrow$ & AUROC$\uparrow$ & AP$\uparrow$\\
            \midrule
            \multirow{2}{*}{Path-based} & Katz Index~\citep{katz1953new} & 0.834 & 0.889 & 0.768 & 0.810 & 0.757 & 0.856\\
            & Personalized PageRank~\citep{page1998pagerank} & 0.845 & 0.899 & 0.762 & 0.814 & 0.763 & 0.860\\
            \midrule
            \multirow{3}{*}{Embeddings} & DeepWalk~\citep{perozzi2014deepwalk} & 0.831 & 0.850 & 0.805 & 0.836 & 0.844 & 0.841\\
            & LINE~\citep{tang2015line} & 0.844 & 0.876 & 0.791 & 0.826 & 0.849 & 0.888\\
            & node2vec~\citep{grover2016node2vec} & 0.872 & 0.879 & 0.838 & 0.868 & 0.891 & 0.914\\
            \midrule
            \multirow{5}{*}{\acp{gnn}} & VGAE~\citep{kipf2016variational} & 0.914 & 0.926 & 0.908 & 0.920 & 0.944 & 0.947\\
            & S-VGAE~\citep{davidson2018hyperspherical} & 0.941 & 0.941 & 0.947 & 0.952 & 0.960 & 0.960\\
            & SEAL~\citep{zhang2018link} & 0.933 & 0.942 & 0.905 & 0.924 & 0.978 & 0.979\\
            & TLC-GNN~\citep{yan2021link} & 0.934 & 0.931 & 0.909 & 0.916 & 0.970 & 0.968\\
            & NBFNet~\citep{zhu2021neural} & 0.956 & 0.962 & 0.923 & 0.936 & 0.983 & 0.982\\
            \midrule
            \acp{snn} & \textbf{\ac{grsnn} (ours)} & 0.936 & 0.945 & 0.915 & 0.931 & 0.982 & 0.982\\
            \bottomrule
        \end{tabular}%
\end{table*}

\subsection{Interpretability}\label{supp:sec:interpretability}

The visualization of the reasoning paths for the final predictions of several examples are shown in \cref{supp:table:interpretability}. It is calculated based on edge and path importance (refer to \cref{supp:sec:experimental_details}). As shown in the results, GRSNN is adept at discerning relation relevances and exploiting transitions, for instance, ``contains'', and analogs, such as individuals with analogous ``award''.

\begin{table*}[t!]
    \centering
    \small
    \caption{\textbf{Visualization of the top-2 reasoning paths for examples on FB15k237.} It is determined by path importances derived from edge importances. The superscript $^{-1}$ indicates the inverse relation.}
    \label{supp:table:interpretability}
        \begin{tabular}{cl}
            \toprule
            \textbf{Query} & $(x, q, y):$ (england, contains, pontefract)\\
            \midrule
            0.967 & (england, contains, west yorkshire) $\land$ (west yorkshire, contains, pontefract)\\
            \hline
            0.671 & (england, contains, leodis) $\land$ (leodis, contains$^{-1}$, west yorkshire) $\land$ (west yorkshire, contains, pontefract)\\
            \midrule
            \textbf{Query} & $(x, q, y):$ (58th academy awards nominees and winners, honored for, kiss of the spider woman (film))\\
            \midrule
            \multirow{2}{*}{1.482} & (58th academy awards nominees and winners, award winner, William Hurt)\\
            & $\land$ (William Hurt, film, kiss of the spider woman (film))\\
            \hline
            \multirow{2}{*}{1.347} & (58th academy awards nominees and winners, award winner, William Hurt)\\
            & $\land$ (William Hurt, nominated for, kiss of the spider woman (film))\\
            \midrule
            \textbf{Query} & $(x, q, y):$ (florida (rapper), profession, artiste)\\
            \midrule
            \multirow{2}{*}{0.513} & (florida (rapper), award, grammy award for album of the year 2010s) \\
            & $\land$ (grammy award for album of the year 2010s, award$^{-1}$, kanye west) $\land$ (kanye west, profession, artiste)\\
            \hline
            \multirow{2}{*}{0.512} & (florida (rapper), award, grammy award for album of the year 2010s) \\
            & $\land$ (grammy award for album of the year 2010s, award$^{-1}$, witney houston) $\land$ (witney houston, profession, artiste)\\
            \bottomrule
        \end{tabular}%
\end{table*}

\begin{figure*}[h!]
    \centering
    \includegraphics[width=0.95\linewidth]{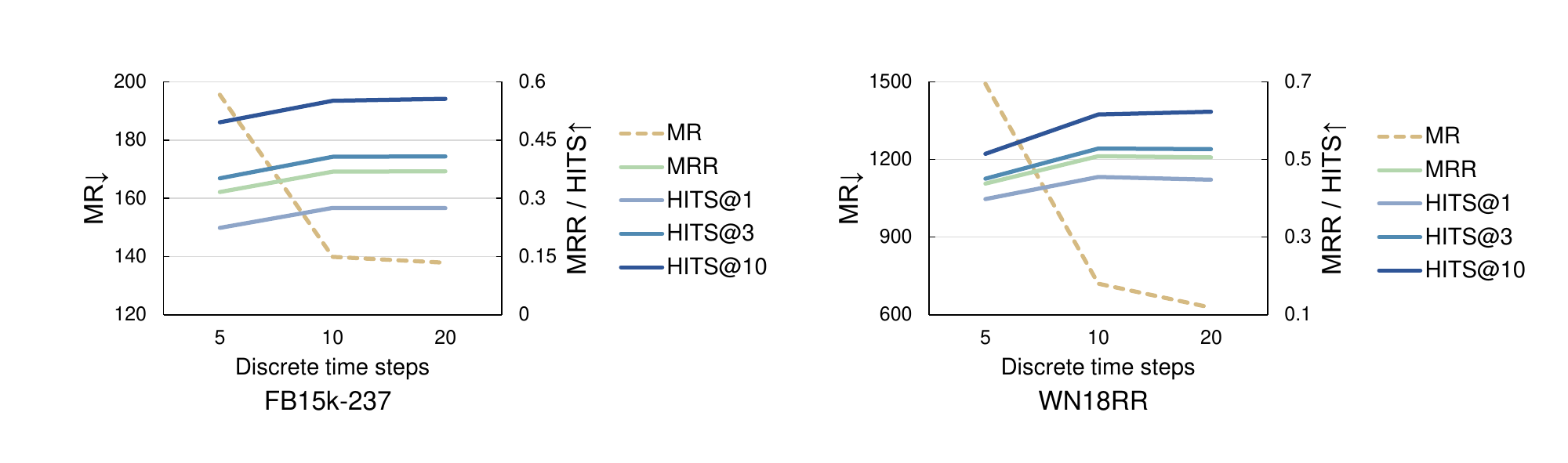}
    \vspace{-4mm}
    \caption{\textbf{Analysis of the temporal discretization of \ac{grsnn} under varying discrete time steps.}}
    \label{supfig:discretization}
\end{figure*}

\subsection{Impact of Discretization Steps}\label{supp:sec:impact_discretization}

The impact of temporal discretization with different intervals and steps on \ac{grsnn} is explored in \cref{supfig:discretization}, and the details of hyperparameters are explained in \cref{supp:sec:experimental_details}. Given the substantial computational cost associated with simulating \acp{snn} over extended periods, experiments primarily employ $T=10$ discrete time steps for \ac{grsnn}. The results indicate that a reduced number of time steps (5) with a larger discretization interval significantly impairs performance due to discretization error, while a larger setting (20) with a smaller interval offers marginal improvements, maintaining comparable results to 10 time steps. This demonstrates the model's robustness under relatively low latency with minimal discrete time steps.

\begin{table}[ht]
    \centering
    \small
    \caption{\textbf{Analysis results of \ac{grsnn} with varying neuron number per graph node on WN18RR.}}
    \label{supp:table:neuron_number}
    \begin{tabular}{ccccccc}
         \toprule
         Neuron number per node & Parameters & MR$\downarrow$ & MRR$\uparrow$ & H@1$\uparrow$ & H@3$\uparrow$ & H@10$\uparrow$\\
         \midrule
         8 & 1.9K & 967 & 0.452 & 0.405 & 0.464 & 0.551\\
         16 & 6.9K & 819 & 0.483 & 0.430 & 0.500 & 0.597\\
         32 & 26K & 720 & 0.508 & 0.455 & 0.528 & 0.616\\
         64 & 101K & 668 & 0.522 & 0.469 & 0.544 & 0.632\\
         96 & 226K & 648 & 0.523 & 0.470 & 0.546 & 0.630\\
         \bottomrule
    \end{tabular}
\end{table}

\begin{table}[t]
    \centering
    \small
    \caption{\textbf{Comparison results of different final \ac{mlp} types on WN18RR.}}
    \label{supp:table:spiking_mlp}
    \begin{tabular}{cccccc}
         \toprule
         Final MLP type & MR$\downarrow$ & MRR$\uparrow$ & H@1$\uparrow$ & H@3$\uparrow$ & H@10$\uparrow$\\
         \midrule
         ReLU & 720 & 0.508 & 0.455 & 0.528 & 0.616\\
         spiking & 701 & 0.502 & 0.447 & 0.521 & 0.615\\
         \bottomrule
    \end{tabular}
\end{table}

\begin{table}[t!]
    \centering
    \small
    \caption{\textbf{Error bar on WN18RR.}}
    \label{supp:table:error_bar}
    \begin{tabular}{ccccc}
         \toprule
         MR$\downarrow$ & MRR$\uparrow$ & H@1$\uparrow$ & H@3$\uparrow$ & H@10$\uparrow$\\
         \midrule
         707$\pm$22 & 0.508$\pm$0.000 & 0.455$\pm$0.001 & 0.528$\pm$0.000 & 0.616$\pm$0.001\\
         \bottomrule
    \end{tabular}
\end{table}

\subsection{Impact of Neuron Number and Parameter Amount}\label{supp:sec:impact_parameter_amount}

As described in \cref{supp:sec:experimental_details}, we take $n=32$ spiking neurons for each graph node, which is consistent with many previous graph neural network works~\citep{zhu2021neural}. To further study the impact of neuron number as well as the corresponding parameter amount, we explore results with varying neuron numbers per node in this section. As shown in \cref{supp:table:neuron_number}, the performance will grow as parameters increase and 32 neurons per node are not optimal. On the other hand, more neurons lead to a larger computational complexity, and there exists a trade-off between performance and complexity.

\subsection{Spiking Link Prediction Head}\label{supp:sec:spiking_head}

As explained in \cref{supp:sec:implementation_details}, we leverage a feedforward neural network $g$ to estimate the conditional likelihood of the given pair representation $\mathbf{h}^q(x, y)$, and following previous work~\citep{zhu2021neural}, $g$ is implemented as a two-layer \ac{mlp} with ReLU activation. While this network can be converted to a spiking \ac{mlp} through rate or temporal coding~\citep{rueckauer2017conversion,stockl2021optimized} to enable a fully spiking system, we further study if we can directly and jointly train a spiking network for this link prediction head. To this end, we replace the ReLU activation with a simple non-leaky Integrate and Fire neuron model combined with the simple current model (\ie, the input current is the linear combination with spikes without dynamics, $I_i(t) = \sum_j w_{ij} s_j(t) + b_i$), consider rate coding over four discrete time steps, and use surrogate derivatives for training. As shown in \cref{supp:table:spiking_mlp}, the performance of spiking \ac{mlp} is almost the same as that with ReLU activation because the major component of the task is to extract pair representations, which is done by the main GRSNN part.

\subsection{Error Bar}

We investigate the error bar of our method on WN18RR based on three runs of experiments with different random seeds. \cref{supp:table:error_bar} shows that the variance of different runs is extremely small, indicating the robustness of the method.

\section{More Discussions}

Knowledge reasoning in human brains involves many neurophysiological processes across many brain areas, and how this is implemented is not fully understood. In this work, we mainly focus on neuro-inspired methods in AI task formulation to investigate how \acp{snn} as computational models can deal with (knowledge) graph reasoning, while it can be future work to study better correspondence with neuroscience.

For neuromorphic hardware, on-chip memory limitation is an issue, so it may not directly support large-scale graphs. But neuromorphic hardware is rapidly developing for larger memories, for example, Intel’s Loihi 2 can support 1 million neurons and 120 million synapses. While this may still not fully support 32 neurons per node for some knowledge graphs, but fewer neurons, such as 16, could be acceptable, and it works for our method with some trade-off for performance (as shown in \cref{supp:sec:impact_parameter_amount}). As neuromorphic computing is a rapidly developing field considering both hardware and algorithm, this paper do not restrict algorithms to some existing hardware, but focuses on the algorithm level with theoretical analysis considering hardware (energy), aligning with previous \acp{snn} works. Actually, algorithms can potentially guide future software-hardware co-design~\citep{schuman2022opportunities}. We hope this work could serve as a catalyst for deeper insights and wider applications of neuromorphic computing systems.


\end{document}